\documentclass[journal]{IEEEtran}

\usepackage{amsmath,amsfonts,amssymb}
\usepackage{tikz,tkz-euclide}
\usetikzlibrary{arrows,calc,shapes,positioning}
\tikzset{>=latex}
\usepackage{color,soul}
\usepackage{graphicx}
\usepackage{calc}
\usepackage{pgfplots}
\usepackage{multirow}
\usepackage{colortbl}

\ifCLASSINFOpdf
\else
\fi
\usepackage{algorithmicx}
\usepackage{algpseudocode}
\ifCLASSOPTIONcompsoc
  \usepackage[caption=false,font=normalsize,labelfont=sf,textfont=sf]{subfig}
\else
  \usepackage[caption=false,font=footnotesize]{subfig}
\fi

\pgfplotsset{compat=1.5}
\begin{document}
\title{IN2LAAMA: INertial Lidar Localisation Autocalibration And MApping}

\author{Cedric Le Gentil${}^\dagger$,~\IEEEmembership{Member,~IEEE,}
        Teresa Vidal-Calleja,~\IEEEmembership{Member,~IEEE,}\\
        and Shoudong Huang,~\IEEEmembership{Senior Member,~IEEE}
\thanks{The authors are with the Centre for Autonomous Systems at the Faculty of Engineering and IT, University of Technology Sydney, Australia.}
\thanks{${}^\dagger$ cedric.legentil@student.uts.edu.au}
\thanks{\textcopyright 2020 IEEE. Personal use of this material is permitted. Permission from IEEE must be obtained for all other uses, in any current or future media, including reprinting/republishing this material for advertising or promotional purposes, creating new collective works, for resale or redistribution to servers or lists, or reuse of any copyrighted component of this work in other works}}

\maketitle

\begin{abstract}
    In this paper, we present INertial Lidar Localisation Autocalibration And MApping (IN2LAAMA): an offline probabilistic framework for localisation, mapping, and extrinsic calibration based on a 3D-lidar and a 6-DoF-IMU.
Most of today's lidars collect geometric information about the surrounding environment by sweeping lasers across their field of view.
Consequently, 3D-points in one lidar scan are acquired at different timestamps.
If the sensor trajectory is not accurately known, the scans are affected by the phenomenon known as motion distortion.
The proposed method leverages preintegration with a continuous representation of the inertial measurements to characterise the system's motion at any point in time.
It enables precise correction of the motion distortion without relying on any explicit motion model.
The system's pose, velocity, biases, and time-shift are estimated via a full batch optimisation that includes automatically generated loop-closure constraints.
The autocalibration and the registration of lidar data rely on planar and edge features matched across pairs of scans.
The performance of the framework is validated through simulated and real-data experiments.
\end{abstract}


\IEEEpeerreviewmaketitle

\section{Introduction}
\definecolor{greentam}{rgb}{0.0, 0.5, 0.0}
\definecolor{bluetam}{rgb}{0.0, 0.0, 0.5}
\definecolor{redtam}{rgb}{0.5, 0.0, 0.0}
\definecolor{orangetam}{rgb}{0.5, 0.25, 0.0}

\def\rc{\mathbf{R}_I^L}
\def\pc{\mathbf{p}_I^L}
\def\lidarpoint{\mathbf{x}_L}
\def\lidarpointi{\lidarpoint^i}
\def\lidarpointj{\lidarpoint^j}
\def\lidarpointk{\lidarpoint^k}
\def\lidarpointl{\lidarpoint^l}
\def\lastlidarpoint{\mathbf{x}_{L_m}}
\def\lastlidarpointi{\lastlidarpoint^i}
\def\imupointi{\mathbf{x}^i_{I}}
\def\worldpoint{\mathbf{x}_{W}}
\def\worldpointi{\worldpoint^i}
\def\worldpointj{\worldpoint^j}
\def\worldpointk{\worldpoint^k}
\def\worldpointl{\worldpoint^l}
\def\time{t}
\def\timei{\time_i}
\def\timek{\time_k}
\def\timekp{\time_{k+1}}
\def\frametime{\tau}
\def\frametimem{\frametime_m}
\def\frametimei{\frametime_i}
\def\frametimemp{\frametime_{m\text{+}1}}
\def\frametimemm{\frametime_{m\text{-}1}}
\def\framesymbol{\mathfrak{F}}
\def\lidarframe{\framesymbol_L}
\def\lidarframei{\lidarframe^{\frametime_i}}
\def\lidarframem{\lidarframe^{\frametimem}}
\def\imuframe{\framesymbol_I}
\def\imuframei{\imuframe^{\timei}}
\def\imuframem{\imuframe^{\frametimem}}
\def\nblidarframe{M}
\def\scanset{\mathcal{X}}
\def\scansetm{\scanset^m}
\def\scansetmm{\scanset^{m-1}}
\def\feature{\mathcal{F}}
\def\featurem{\feature^m}
\def\featuremp{\feature^{m+1}}
\def\association{\mathcal{A}}
\def\oneassociation{\mathfrak{a}}
\def\acc{\mathbf{f}}
\def\gyr{\boldsymbol{\omega}}
\def\acci{\acc_{i}}
\def\gyri{\gyr_{i}}
\def\imutimei{\mathfrak{t}_i}
\def\imutime{\mathfrak{t}}
\def\acct{{\acc}^*(t)}
\def\gyrt{{\gyr}^*(t)}
\def\imurot{\mathbf{R}_W}
\def\imupos{\mathbf{p}_W}
\def\imuvel{\mathbf{v}_W}
\def\imurotTi{{\imurot^{\frametimei}}}
\def\imurotm{{\imurot^{\frametimem}}}
\def\imuposm{{\imupos^{\frametimem}}}
\def\imuvelm{{\imuvel^{\frametimem}}}
\def\imurotmp{{\imurot^{\frametimemp}}}
\def\imuposmp{{\imupos^{\frametimemp}}}
\def\imuvelmp{{\imuvel^{\frametimemp}}}
\def\imurotmm{{\imurot^{\frametimemm}}}
\def\imuposmm{{\imupos^{\frametimemm}}}
\def\imuvelmm{{\imuvel^{\frametimemm}}}
\def\imuroti{{\imurot^{\timei}}}
\def\imuposi{{\imupos^{\timei}}}
\def\imuveli{{\imuvel^{\timei}}}
\def\bacc{\mathbf{b}_f}
\def\baccm{\bacc^m}
\def\baccmm{\bacc^{m-1}}
\def\baccpriorm{\bar{\mathbf{b}}_f^{m}}
\def\baccpriormm{\bar{\mathbf{b}}_f^{m-1}}
\def\bacccorrectionz{\hat{\mathbf{b}}_f^{0}}
\def\bacccorrectionm{\hat{\mathbf{b}}_f^{m}}
\def\bacccorrectionM{\hat{\mathbf{b}}_f^{M\text{-}1}}
\def\bacccorrectionmm{\hat{\mathbf{b}}_f^{m-1}}
\def\bgyr{\mathbf{b}_\omega}
\def\bgyrm{\bgyr^m}
\def\bgyrmm{\bgyr^{m-1}}
\def\bgyrpriorm{\bar{\mathbf{b}}_\omega^{m}}
\def\bgyrpriormm{\bar{\mathbf{b}}_\omega^{m-1}}
\def\bgyrcorrectionz{\hat{\mathbf{b}}_\omega^{0}}
\def\bgyrcorrectionm{\hat{\mathbf{b}}_\omega^{m}}
\def\bgyrcorrectionM{\hat{\mathbf{b}}_\omega^{M\text{-}1}}
\def\bgyrcorrectionmm{\hat{\mathbf{b}}_\omega^{m-1}}
\def\timeshift{\delta_t}
\def\timeshiftm{\timeshift^m}
\def\timeshiftpriorm{\bar{\delta}_t^{m}}
\def\timeshiftcorrectionz{\hat{\delta}_t^{0}}
\def\timeshiftcorrectionm{\hat{\delta}_t^{m}}
\def\timeshiftcorrectionM{\hat{\delta}_t^{M\text{-}1}}
\def\worldframe{\mathfrak{F}_W}
\def\state{\mathbf{\mathcal{S}}}
\def\mth{m^{\textit{th}}}
\def\measurements{\mathbf{\mathcal{Z}}}
\def\lidarresidual{d}
\def\lidarresiduala{\lidarresidual_{\oneassociation}}
\def\imuresidual{\mathbf{r}_I}
\def\imuresidualm{\imuresidual^m}
\def\imurresidualm{\mathbf{r}_{I_r}^m}
\def\imuvresidualm{\mathbf{r}_{I_v}^m}
\def\imupresidualm{\mathbf{r}_{I_p}^m}
\def\baccresidual{\mathbf{r}_f}
\def\bgyrresidual{\mathbf{r}_{\omega}}
\def\timeshiftresidual{r_t}
\def\baccresidualm{\baccresidual^m}
\def\bgyrresidualm{\bgyrresidual^m}
\def\timeshiftresidualm{\timeshiftresidual^m}
\def\upmrot{\Delta\mathbf{R}_{\frametimem}}
\def\upmpos{\Delta\mathbf{p}_{\frametimem}}
\def\upmvel{\Delta\mathbf{v}_{\frametimem}}
\def\upmrotmp{\upmrot^{\frametimemp}}
\def\upmposmp{\upmpos^{\frametimemp}}
\def\upmvelmp{\upmvel^{\frametimemp}}
\def\upmrotmm{\Delta\mathbf{R}^{\frametimem}_{\frametimemm}}
\def\upmposmm{\Delta\mathbf{p}^{\frametimem}_{\frametimemm}}
\def\upmvelmm{\Delta\mathbf{v}^{\frametimem}_{\frametimemm}}
\def\upmroti{\upmrot^{\timei}}
\def\upmposi{\upmpos^{\timei}}
\def\upmveli{\upmvel^{\timei}}
\def\gravity{\mathbf{g}}
\def\Tc{\mathbf{T}_I^L}
\def\imuT{\mathbf{T}_W}
\def\imuTi{{\imuT^{t_i}}}
\def\imuTm{{\imuT^{\frametimem}}}
\def\imuTTi{{\imuT^{\frametimei}}}
\def\identitymat{\mathbf{I}}

\begin{figure}
    \centering
    
    \def\imageheight{4}
    \def\imagesmallheight{3.8}
    \def\imagerest{0.2}
    \def\imagespread{2}
    \newcommand\sensorsuit[3]{
        \node[anchor=south west] at (#1+0,#2+0) {\includegraphics[clip, height=#3cm]{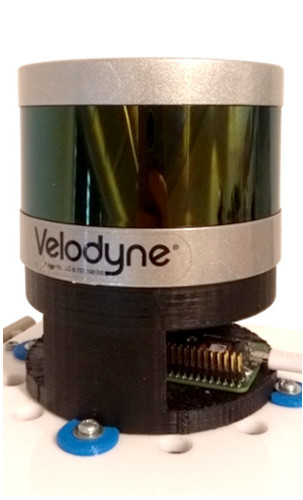}};
        \draw[red, thick,rounded corners=0.5] (0.06*#3+#1,0.45*#3+#2) rectangle (0.61*#3+#1,0.93*#3+#2);
		\draw[red, thick,rounded corners=0.5] (0.59*#3+#1,0.35*#3+#2) -- (0.53*#3+#1,0.25*#3+#2) -- (0.28*#3+#1,0.29*#3+#2) -- (0.37*#3+#1,0.38*#3+#2) -- cycle;
		\node[anchor=south west, red] at (0.04*#3+#1,0.91*#3+#2) {\scriptsize $Lidar$};
        \node[anchor=north west, red] at (0.41*#3+#1,0.20*#3+#2) {\scriptsize $I\!M\!U$};
    }
    \begin{tikzpicture}
        \node[anchor=south west] at (0*\imagespread,\imagerest) {\includegraphics[clip, height=\imagesmallheight cm]{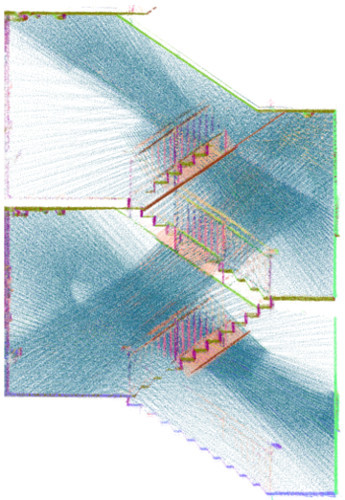}};
        \node[anchor=south west] at (1.9*\imagespread,\imagerest) {\includegraphics[clip, height=\imagesmallheight cm]{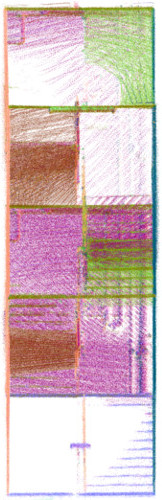}};
        \sensorsuit{3*\imagespread}{0}{\imageheight}
        \node[anchor=south west, text width= 6cm, execute at begin node=\setlength{\baselineskip}{4pt}] at (0.1,0) {\tiny Coloured with post-processed normals (100-closest neighbours)};
	\end{tikzpicture}
    \caption{Left and centre: Map generated with IN2LAAMA in a staircase. Right: sensor suite used (Velodyne VLP-16 lidar and Xsens MTi-3 IMU).}
    \label{figure:teaser}
\end{figure}
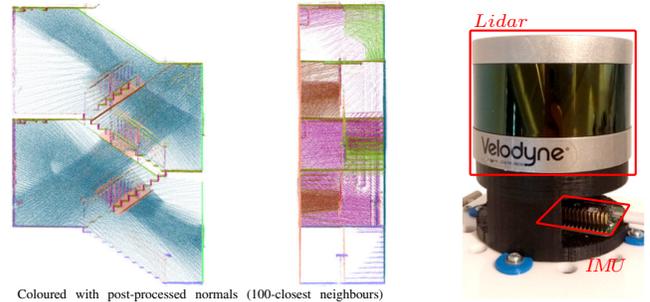

\IEEEPARstart{L}{ocalisation} and mapping is a key component of any autonomous system operating in unknown or partially known environments.
In a world that relies more and more on automation and robotics, the need for 3D models of the environment is increasing at a fast pace.
Autonomous systems navigation is not the only source of demand for 3D maps.
A growing number of fields are gaining interest in dense and accurate representations, especially for monitoring or inspection operations, and augmented reality purposes.
This manuscript presents a probabilistic framework for localisation and mapping with targetless extrinsic calibration that tightly integrates data from a 3D-lidar range scanner and a 6-DoF-Inertial Measurement Unit (IMU).
We have named this approach \emph{INertial Lidar Localisation Autocalibration And MApping} (IN2LAAMA).
Fig.~\ref{figure:teaser} shows a map example generated with data collected with our sensor system.

In its early days, as shown in \cite{DurrantWhyte1996}, localisation of autonomous vehicles was mainly relying on the knowledge of the position of beacons in the environment.
The same concept is used by GPS, where multiple satellites orbiting more than twenty thousands kilometres above our heads serve as beacons.
While GPS provides substantial localisation information in many outdoor environments worldwide, it cannot be used indoors or in scenarios where line-of-sight with the satellites cannot be guaranteed.
The non-stoping evolution of lidar technologies in the past decades permitted great advancements in the simultaneous localisation and mapping field.
Despite providing crucial information about real-world geometry, today's 3D-lidars still suffer from some drawbacks.
Due to its sweeping mechanism (be it with the help of a spinning mechanism, actioned mirrors, prisms, etc.) a lidar does not take a snapshot of the environment but progressively scans the surrounding space.
Accurate knowledge of the motion of the sensor during the sweeps is needed to allow the grouping of the collected 3D-points in consistent scans.
Inaccuracies in the trajectory will introduce \textit{motion distortion} in the resulting point cloud.

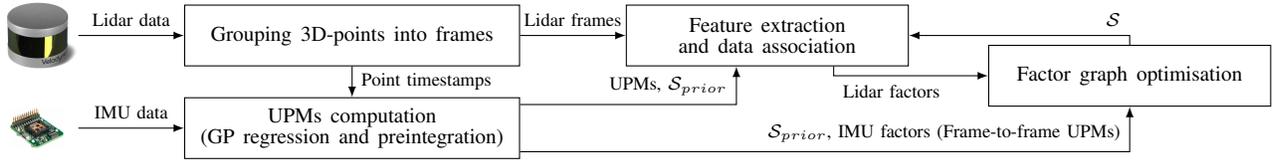
\begin{figure*}
    \centering
    \begin{tikzpicture}[auto]
        \tikzstyle{block} = [draw, fill=white, rectangle, minimum height = 2.3em, text width = 10em,  minimum width = 10em, text centered, node distance = 5em, execute at begin node=\setlength{\baselineskip}{8pt}]
        \def\inputdist{4em}
        \def\otherdist{2.8em}

        \node (lidar_input){};
        \node [block, right = \inputdist of lidar_input, text width = 12em] (framefactory){\footnotesize Grouping 3D-points into frames};
        \node [below= \otherdist of lidar_input] (imu_input){};
        \node [block, right = \inputdist of imu_input, text width = 12em] (upm){\footnotesize UPMs computation\\(GP regression and preintegration)};
        \node [block, right = \inputdist of framefactory] (feature_extraction){\footnotesize Feature extraction and data association};
        \node [below = 0.0em of feature_extraction] (midpoint){};
        \node [block, right = 2*\inputdist of midpoint] (optimisation){\footnotesize Factor graph optimisation};

        \node[anchor = center, left = -1em of lidar_input](lidar_image) {\includegraphics[clip, trim=0cm 0cm 0cm 0cm, height=2.5em]{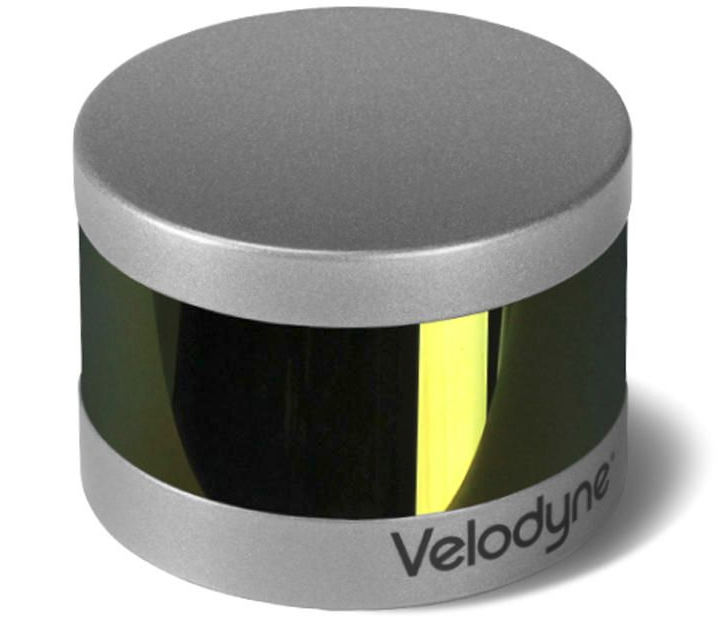}};
        \node[anchor = center, left = -0.5em of imu_input](imu_image) {\includegraphics[clip, trim=0cm 0cm 0cm 0cm, height=1.5em]{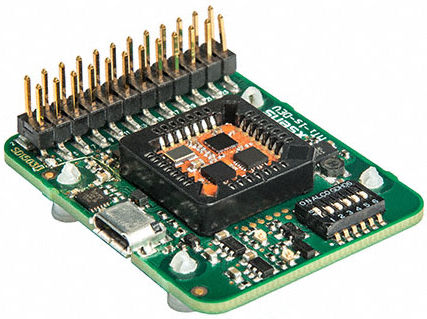}};

        \draw[->] (lidar_input) -- node {\scriptsize Lidar data} (framefactory);
        \draw[->] (imu_input) -- node {\scriptsize IMU data} (upm);
        \draw[->] (framefactory.east) -- node {\scriptsize Lidar frames} (feature_extraction.west);
        \draw[->] (framefactory.south) -- node {\scriptsize Point timestamps} (framefactory.south|-upm.north);
        \draw[->] (upm.8) -|  node[anchor=south east] {\scriptsize UPMs, $\state_{prior}$} (feature_extraction.225);
        \draw[->] (upm.352) -| node [anchor=south east] {\scriptsize $\state_{prior}$, IMU factors (Frame-to-frame UPMs)} (optimisation.south);
        \draw[->] (feature_extraction.335) |- node [anchor=north west] {\scriptsize Lidar factors} (optimisation.west);
        \draw[->] (optimisation.north) |- node [anchor=south east] {\scriptsize $\state$} (feature_extraction.east);

    \end{tikzpicture}
    \captionsetup[subfigure]{justification=centering}
	\caption{
        Overview of IN2LAAMA with $\state$ the estimated state.
	}
	\label{figure:overview}
\end{figure*}

By leveraging inertial data, the proposed method aims to accurately correct motion distortion in lidar scans without making explicit assumptions about the system's motion.
In this work, as introduced in \cite{LeGentil2018}, the concept of preintegration (\cite{Lupton2012}, \cite{Forster2015a}) is used over interpolated IMU measurements.
These generated pseudo-measurements are named Upsampled Preintegrated Measurements (UPMs) and allow the generation of inertial data at any time.
Gaussian Process (GP) inference is used for non-parametric probabilistic interpolation.
UPMs inherit all the properties from the original preintegration \cite{Lupton2012}, which include the independence of the measurements with respect to the initial pose and velocity conditions.

Like most localisation and mapping frameworks, the proposed, method is constituted of two main modules: a front-end for feature extraction and data association, and a back-end for state estimation through numerical optimisation.
Nonetheless, IN2LAAMA differs from most frameworks as per the tight relationship between these two major modules.
Reliable geometric feature extraction in lidar scans requires the knowledge of the system's trajectory to be unaffected by motion distortion.
Accurate knowledge of the system's trajectory relies on the extraction and association of robust features.
To address this ``chicken-and-egg" problem, the features (front-end) are periodically recomputed according to the last state estimate (back-end), as shown in the block diagram of Fig.~\ref{figure:overview}.

Our work does not aim at real-time operation but focuses on accurate 3D mapping given no prior knowledge.
Therefore, an offline batch optimisation framework is proposed here.
This paper extends our previous work on lidar-inertial localisation and mapping \cite{LeGentil2019}.
The first key contribution with respect to our previous work is the use of IMU factors between consecutive poses and velocities of the estimated state.
These additional constraints in the optimisation generalise the full formulation of lidar-inertial integration for localisation and mapping and provide more accurate results.
This paper also contains an analysis of the impact that poorly modelled inertial data have on the accuracy of the proposed method.
The second important contribution of this work is the ability to perform targetless extrinsic calibration between a 3D-lidar and a 6-DoF IMU simultaneously to localisation and mapping.
A simple loop-closure detection method and better outlier rejection strategies have been integrated into the proposed framework.

The remainder of the paper is organised as follows.
Section~II discusses related work.
The method overview is given in Section~III.
The back-end of IN2LAAMA is explained in Section~IV.
Section~V details the front-end of the method.
In Section~VI, we explain implementation details and the strategy employed for building our optimisation approach.
Section~VII presents simulated and real-world experimental results.
This paper comes along a video that shows 3D animations of the results in real-world scenarios.
Finally, conclusion and future work are discussed in Section~VIII.

\section{Related work}

Lidar scan geometric registration is the foundation of most of the laser-based localisation algorithms.
The introduction of Iterative Closest Points (ICP) \cite{Besl1992} and generalised ICP \cite{Segal2009} allowed the estimation of the rigid transformation between two point clouds.
The method in \cite{Mendes2016} estimates a system trajectory using ICP for frame-to-frame relative poses estimation, and a pose-graph to correct the drift inherent to odometry-like frameworks.
This method, among others, does not address the phenomenon of motion distortion in lidar scans.

Different approaches have been presented in the literature for state estimation of moving lidar/rolling-shutter sensors.
The authors of \cite{Hong2010} extended the 2D standard ICP to account for the lidar motion using the assumption of constant velocity during the sweep.
This motion model assumption is often used in the literature as in \cite{Zhang2014b} and \cite{Bosse2009}.
Both techniques rely on linear pose-interpolation between control points.
In many real-world scenarios, these motion model assumptions do not represent the true nature of the system's motion.

In \cite{Furgale2012}, the authors use a linear combination of temporal basis functions to represent the state.
While providing greater representability compared to traditional discrete models, the performances of continuous state frameworks depend on the veracity of the models assumed.
Nonetheless, continuous pose representation, as introduced in \cite{Bosse2009} or \cite{Furgale2012}, allows the use of non-synchronised sensors in multi-modal frameworks.

A probabilistic approach for continuous state estimation is presented in \cite{Anderson2015}.
It uses computationally efficient GP regression over a discrete maximum a posteriori estimation to allow continuous inference of the state variables. 
Nonetheless, this method still relies on large discrete state estimation.

To reach real-time operations, techniques like \cite{Zhang2014b} and \cite{Droeschel2018} consider the correction of motion distortion at the front-end level, and not as part of the estimation problem.
In other words, the prior knowledge of the actual motion is used to undistort the incoming point clouds, but no other action is conducted later to improve the distortion correction according to the new state estimate.
With such a strategy, there is a risk of accumulating drift due to inaccurate initial conditions.
Our proposed framework constantly revisits the motion distortion correction all along the estimation process.

While unreliable for long-run motion estimation on their own, IMUs have been extensively used in combination with exteroceptive sensors like cameras and lidars to develop robust multi-modal systems.
Initially proposed for visual-inertial fusion in \cite{Lupton2012} and \cite{Forster2015a}, the concept of preintegration allows the pre-processing of inertial measurements for state estimation.
The key idea is to dissociate the acceleration integration from the initial pose and velocity.
By doing so, the inertial integrations can be computed beforehand, and the resulting pseudo-measurements stay constant during the estimation process, regardless of the linearisation point changes.
In our earlier work on lidar-IMU extrinsic calibration \cite{LeGentil2018}, we extended this concept by considering preintegration over continuous representation of the inertial data.
The UPMs allow the computation of inertial data at any timestamp and therefore, enable the characterisation of the system's motion during a lidar scan without the use of any explicit motion or state model.
The localisation and mapping work presented here leverages the UPMs in a graph-based optimisation framework that does not rely on any explicit motion model.

The 3D-maps generated in \cite{Bosse2012} and \cite{Park2018} make use of surfels.
Given dense enough 3D-point clouds, surfels can provide rich surface information.
Various front-end solutions have been explored in the literature to deal with the sparsity of 3D-data generated by nowadays lidars.
The approach presented in \cite{Geneva2018} estimates both the system trajectory and the position of planes in the environment based on a novel plane representation.
Such formulation reduces the complexity of the optimisation problem and filters the individual measurements' noise. But this method is adapted solely to highly structured environments.
The authors of \cite{Serafin2016} proposed a lidar feature extraction method that efficiently detects planes and 3D-lines in structured environments.
Using a specific sampling of lidar scans, IMLS-SLAM \cite{Deschaud2018} leverage more generic ``features", and can be used in weakly structured environments.
The per-channel feature extraction in LOAM \cite{Zhang2014b} does not require the computation of surface information (normals) for each of the lidar points.
Consequently, this approach is suitable for lidars with lower resolution and inspired our front-end development.

Most of the localisation and mapping techniques mentioned above rely on known extrinsic calibration between the different sensors of the system.
Despite the popularity of lidars and IMUs, the calibration between these sensors has not been extensively studied.
Our previous work \cite{LeGentil2018} proposed a calibration process based on a simple calibration target (a set of planes).
As a general rule, the use of calibration targets allows the calibration frameworks to leverage prior knowledge of the data observed during the calibration recording.
For example, the visual-inertial calibration method presented in \cite{Furgale2013} uses a known checkerboard to determine the camera position.
Few problems can arise from the fact of using a calibration target.
The first one is the manufacture of the target itself.
An imprecise target leads to inaccurate calibration parameters.
The second, and maybe the most important issue in an industrial/commercial context, is the need for a dedicated calibration procedure with a specific calibration rig.
In other words, it means that a user, or often a qualified operator, needs to perform a series of pre-defined actions to re-calibrate the system.
Work has been conducted to move toward targetless calibration procedures in the case of lidar-visual calibration (\cite{Taylor2014} \cite{Castorena2016}).
Our proposed method follows this line of thought by allowing targetless extrinsic calibration of a 3D-lidar and a 6-DoF-IMU, thus removing the need for a dedicated calibration rig/environment/pre-defined actions.

\section{Method overview}

\subsection{Notation and definitions}

Let us consider a rigidly mounted 3D lidar and a 6-DoF IMU.
The lidar and IMU reference frames at time $\timei$ are noted $\lidarframe^{\timei}$ and $\imuframe^{\timei}$ respectively.
The rotation matrix $\rc$ and the translation vector $\pc$ characterise the pose of $\lidarframe^{\timei}$ in $\imuframe^{\timei}$.
Homogeneous transformation will be used for the rest of the paper, therefore rotation matrices and translations/positions will be associated with $4\times4$ transformation matrices with the same combination of subscripts and superscripts,
\begin{equation}
    \mathbf{T}_a^b = \begin{bmatrix}\mathbf{R}_a^b&\mathbf{p}_a^b\\\mathbf{0}^\top & 1\end{bmatrix} \text{ and }
        {\mathbf{T}_a^b}^{-1} = \begin{bmatrix}{\mathbf{R}_a^b}^\top&-{\mathbf{R}_a^b}^\top\mathbf{p}_a^b\\\mathbf{0}^\top & 1\end{bmatrix}.
\end{equation}
The 3D-points $\lidarpointi$ provided by the lidar at time $\timei$ are projected from $\lidarframe^{\timei}$ to $\imuframe^{\timei}$ using 
\begin{equation}
	\begin{bmatrix}\imupointi\\1\end{bmatrix} = \Tc \begin{bmatrix}\lidarpointi\\1\end{bmatrix}.
	\label{eq:calib_projection}
\end{equation}

In this work, the lidar points are grouped into $\nblidarframe$ frames.
Note that in the proposed method, a frame corresponds to the data collected in scan greater than 360-degree, as explained in Section~V-C. 
The points that belong to the $m^{\textit{th}}$ frame form the set $\scansetm$.
$\featurem$ is a subset of $\scansetm$ that represents lidar feature-points.
A feature is a point belonging to a distinctive type of surface (e.g. plane or edge).
The set of feature associations $\association$ contains tuples of 3 or 4 lidar feature-points depending on whether they are edges or planes respectively.

The 6-DoF-IMU is the combination of a 3-axis accelerometer and a 3-axis gyroscope.
Therefore, the inertial data acquired consists of proper accelerations $\acci$ and angular velocities $\gyri$ at time $\imutimei$ ($i=1,\dots,Q$).
GP regression is used to infer inertial readings on each IMU DoF independently at any given time $t$.
The continuous readings $\acct$ and $\gyrt$, estimated using GPs, allow the attribution of IMU readings to each of the individual lidar points.

The proposed method aims to estimate the IMU orientation $\imurotm$, position $\imuposm$ and velocity $\imuvelm$ for each lidar frame ($m = 0, \dots,M\text{-}1$), as well as the IMU biases and the time-shifts between the two sensors.
The subscript $_W$ represents the earth-fixed world reference frame $\worldframe$, and $\frametimem$ corresponds to the timestamp at the beginning of the $\mth$ lidar frame.
$\framesymbol_\bullet^{\frametimem}$ refers to the reference frame of the IMU or lidar (as $\bullet$ represents in this case $L$ or $I$) at time $\frametimem$.

In the following, $\state$ indicates the state to be estimated:
$\state = (\imurot^{\frametime_0},\cdots,\imurot^{\frametime_{M\text{-}1}}$, $\imupos^{\frametime_1},\cdots,\imupos^{\frametime_{M\text{-}1}}$, $\imuvel^{\frametime_0},\cdots,\imuvel^{\frametime_{M\text{-}1}}$, $\bacccorrectionz,\cdots,\bacccorrectionM$, $\bgyrcorrectionz, \cdots,\bgyrcorrectionM$, $\timeshiftcorrectionz, \cdots, \timeshiftcorrectionM)$ with $\bacccorrectionm$, $\bgyrcorrectionm$, and $\timeshiftcorrectionm$ the biases and time-shift corrections associated with the $\mth$ lidar frame (more details about the biases and time-shift corrections are given in Section~IV-C).
The calibration procedure, explained in Section~VI.B, adds the calibration parameters $\Tc$ to the state $\state$.

\subsection{Cost function}

The proposed method does not rely on any trajectory prior but uses a Gaussian distribution to constrain the inter-sensor time-shift.
Therefore, the localisation and mapping problem is formulated as a Maximum A Posterior (MAP) estimation:
\begin{equation}
\state^* =
\underset{\state}{\text{argmin}}\;  -\log(p(\measurements|\state)p(\state))
= \underset{\state}{\text{argmin}}\;  C{(\state)},
\end{equation}
with $\measurements$ representing the available measurements and $C$ the optimisation cost function.

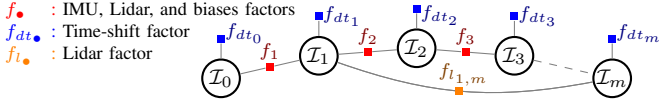
\begin{figure}

    \def\framenode{\mathcal{I}}
    \def\lidarfactor{l}

    \centering
	\begin{tikzpicture}

    \def\posxscale{0.65}
    \def\xa{\posxscale*0}
    \def\xb{\posxscale*2}
    \def\xc{\posxscale*4}
    \def\xd{\posxscale*6}
    \def\xm{\posxscale*8}
    \def\ya{-0.4}
    \def\yb{-0.1}
    \def\yc{0}
    \def\yd{-0.1}
    \def\ym{-0.4}
   
    \def\dtdist{0.2}
    \def\residualdist{-0.05}
    
    \def\xl{-3.0}
    \def\yl{-0.1}
    \def\interline{0.3}

    \begin{scope}[every node/.style={circle, thick,draw}]
        \node[minimum size=0.5cm, inner sep=0pt] (I0) at (\xa,\ya) {\footnotesize $\framenode_0$};
        \node[minimum size=0.5cm, inner sep=0pt] (I1) at (\xb,\yb) {\footnotesize $\framenode_1$};
        \node[minimum size=0.5cm, inner sep=0pt] (I2) at (\xc,\yc) {\footnotesize $\framenode_2$};
        \node[minimum size=0.5cm, inner sep=0pt] (I3) at (\xd,\yd) {\footnotesize $\framenode_3$};
        \node[minimum size=0.5cm, inner sep=0pt] (Im) at (\xm,\ym) {\footnotesize $\framenode_m$};
    \end{scope}

    \begin{scope}[every node/.style={fill,gray}]
        \node[inner sep=1.5pt,red] (f01) at (\xa*0.5+\xb*0.5,\ya*0.5+\yb*0.5){};
        \node[inner sep=1.5pt,red] (f12) at (\xb*0.5+\xc*0.5,\yb*0.5+\yc*0.5){};
        \node[inner sep=1.5pt,red] (f23) at (\xc*0.5+\xd*0.5,\yc*0.5+\yd*0.5){};
        \node[inner sep=1.5pt,blue, above = \dtdist of I0] (f0){};
        \node[inner sep=1.5pt,blue, above = \dtdist of I1] (f1){};
        \node[inner sep=1.5pt,blue, above = \dtdist of I2] (f2){};
        \node[inner sep=1.5pt,blue, above = \dtdist of I3] (f3){};
        \node[inner sep=1.5pt,blue, above = \dtdist of Im] (fm){};
    \end{scope}

    \begin{scope}[auto,gray]
        \draw (I0) -- (f01) -- (I1) -- (f12) -- (I2) -- (f23) -- (I3);
        \path (Im) edge [bend left=15] node[orangetam, anchor=south, yshift=-0.04cm]{\scriptsize$f_{\lidarfactor_{1,m}}$} (I1);
        \path (Im) edge [bend left=15] node[fill,inner sep=1.5pt,orange,yshift=0.06cm]{} (I1);
        \draw (I0) -- (f0);
        \draw (I1) -- (f1);
        \draw (I2) -- (f2);
        \draw (I3) -- (f3);
        \draw (Im) -- (fm);
        \draw [dashed](I3) -- (Im);
    \end{scope}

    \begin{scope}[auto]
        \node [bluetam, right = \residualdist of f0,anchor=west, xshift=-0.08cm,yshift=0.08cm] {\scriptsize$f_{dt_0}$};
        \node [bluetam, right = \residualdist of f1,anchor=west, xshift=-0.08cm,yshift=0.00cm] {\scriptsize$f_{dt_1}$};
        \node [bluetam, right = \residualdist of f2,anchor=west, xshift=-0.08cm,yshift=-0.08cm] {\scriptsize$f_{dt_2}$};
        \node [bluetam, right = \residualdist of f3,anchor=west, xshift=-0.08cm,yshift=0.00cm] {\scriptsize$f_{dt_3}$};
        \node [bluetam, right = \residualdist of fm,anchor=west, xshift=-0.08cm,yshift=0.08cm] {\scriptsize$f_{dt_m}$};
        \node [redtam,above = \residualdist of f01, anchor=south, yshift=-0.04cm]{\scriptsize$f_{1}$};
        \node [redtam,above = \residualdist of f12, anchor=south, yshift=-0.04cm]{\scriptsize$f_{2}$};
        \node [redtam,above = \residualdist of f23, anchor=south, yshift=-0.04cm]{\scriptsize$f_{3}$};
    \end{scope}

    \begin{scope}[auto]
        \node[anchor = west] at (\xl,\yl) {\scriptsize\textcolor{orange}{$f_{l_\bullet}\;\:$ :} Lidar factor};
        \node[anchor = west] at (\xl,\yl+\interline) {\scriptsize\textcolor{blue}{$f_{dt_\bullet}$ :} Time-shift factor};
        \node[anchor = west] at (\xl,\yl+2*\interline) {\scriptsize\textcolor{red}{$f_{\bullet}\ \;\;$ :} IMU, Lidar, and biases factors};
    \end{scope}

    \end{tikzpicture}
	\caption{
		Factor graph representation of the optimisation problem solved in IN2LAAMA.
        $\framenode_m$ = \{$\imurotm$, $\imuposm$, $\imuvelm$, $\bacccorrectionm$, $\bgyrcorrectionm$, $\timeshiftcorrectionm$\} represents the IMU pose, velocity, biases and time-shift correction associated to the lidar scan $\scansetm$ at $\frametimem$. The factor $f_{\lidarfactor_{2,m}}$ represents a loop-closure.
	}
	\label{figure:factor_graph}
\end{figure}

Represented as a factor graph in Fig.~\ref{figure:factor_graph}, and under the assumption of zero-mean Gaussian noise, the estimation can be solved by minimising geometric distances $d_\oneassociation$ associated with lidar features, inertial residuals $\imuresidualm$, accelerometer biases residuals $\baccresidualm$, gyroscope biases residuals $\bgyrresidualm$, and time-shift residuals $\timeshiftresidualm$. That is,
\begin{align}
C{(\state)} =  &\sum_{\oneassociation\in\association} \lVert \lidarresiduala \lVert_{\Sigma_{\lidarresiduala}}^2  + \sum_{m=0}^{M-1} \lVert \timeshiftresidualm \lVert_{\Sigma_{\timeshiftresidualm}}^2  +
\nonumber
\\
&\sum_{m=1}^{M-1} \Big( \lVert \baccresidualm \lVert_{\Sigma_{\baccresidualm}}^2 + \lVert \bgyrresidualm \lVert_{\Sigma_{\bgyrresidualm}}^2 + \lVert \imuresidualm \lVert_{\Sigma_{\imuresidualm}}^2 \Big).
\label{eq:optimisation}
\end{align}
The different components of $C(\state)$ are detailed in Section~IV.
Note that $\Sigma_\bullet$ is the covariance matrix of the variable $\bullet$.

\subsection{Upsampled Preintegrated Measurement}

The proposed framework uses UPMs to address the  problem of motion distortion in lidar scans accurately.
We previously introduced these measurements in \cite{LeGentil2018} based on concepts originally presented in \cite{Lupton2012} and \cite{Forster2015a}.
The original idea of preintegration \cite{Lupton2012} consists of reducing the size of the state to estimate by combining IMU measurements between two estimated poses.
From acceleration and angular velocities, IMU readings are naturally combined through integration.
The problem is, as per the physics definition of accelerometer readings, the integration is computed based on initial conditions, which become part of the estimated state.
In such a configuration, every modification of the state would require recomputation of all the integrals.
Preintegration allows for the computation of the integrals independently from the initial conditions.
In other words, authors of \cite{Forster2015a} created a new type of measurements that remain constant during the estimation process and that links two consecutive poses of the state. 

The original preintegrated measurements, as defined in \cite{Forster2015a}, are 
\begin{align}
&\begin{aligned}
    \upmposi\! =& \textstyle\sum_{k = \kappa}^{i-1}\Big(\upmvel^{\timek}\Delta \timek\!+\!\upmrot^{\timek}(\mathbf{f}(\timek\!-\!\timeshiftm)\!-\!\baccm)\textstyle\frac{{\Delta \timek}^2}{2}\Big)
\\
\upmveli\! =& \textstyle\sum_{k = \kappa}^{i-1} \upmrot^{\timek}(\mathbf{f}(\timek\!-\!\timeshiftm)\!-\!\baccm)\Delta \timek
\nonumber
\end{aligned}
\\
&\begin{aligned}
\upmroti\! = \textstyle\prod_{k = \kappa}^{i-1} \text{Exp}\big((\boldsymbol{\omega}(\timek\!-\!\timeshiftm)\!-\!\bgyrm)\Delta \timek\big),
\end{aligned}
\label{eq:imu_delta}
\end{align}
with $\{\kappa \in \mathbb{N} | t_\kappa = \tau_m\}$, $\Delta \timek = \timekp - \timek$, and $\text{Exp}(.)$ the exponential mapping from the axis-angle representations ($\mathfrak{so}(3)$) to rotation matrices ($SO(3)$) defined as
\begin{align}
    &\text{Exp}\big(\boldsymbol{\phi}) = \identitymat + \textstyle\frac{\sin(\lVert \boldsymbol{\phi} \lVert)}{\lVert \boldsymbol{\phi} \lVert} \boldsymbol{\phi}^\wedge + \textstyle\frac{1-\cos(\lVert \boldsymbol{\phi} \lVert)}{\lVert \boldsymbol{\phi} \lVert^2} \big(\boldsymbol{\phi}^\wedge\big)^2,\\
    &\text{where } \boldsymbol{\phi}^{\wedge} = \begin{bmatrix} \phi_1 \\ \phi_2 \\ \phi_3 \end{bmatrix}^{\wedge} = \begin{bmatrix} 0 & -\phi_3 & \phi_2 \\ \phi_3 & 0 & -\phi_1 \\ -\phi_2 & \phi_1 & 0 \end{bmatrix}.
\end{align}
Then the pose and velocity of the IMU at time $\timei$ are
\begin{align}
\imuposi &= \imuposm + \Delta \varsigma_m^i \imuvelm + \frac{1}{2}{\Delta \varsigma_m^i}^2\gravity + \imurotm\upmposi
\label{eq:imu_p}
\\
\imuveli &= \imuvelm + \Delta \varsigma_m^i\gravity + \imurotm\upmveli
\label{eq:imu_v}
\\
\imuroti &= \imurotm\upmroti
\label{eq:imu_r},
\end{align}
where $\gravity$ is the known gravity vector in $\worldframe$ and $\Delta \varsigma_m^i = \timei - \frametimem$.

The issue addressed by the UPMs is the general asynchronism between the IMU and any other sensor, both in terms of time-shift and difference of acquisition frequency.
For the lidar-IMU pair, unlike in \cite{Lupton2012} and \cite{Forster2015a}, the IMU is the low acquisition frequency sensor.
With a rotating lidar moving in space, each of the points is collected from a different pose.
To constrain the motion during a sweep, inertial data need to be available at each lidar point's timestamp.
The key idea of UPMs is to compute a continuous representation of the inertial data to allow for its estimation at any arbitrary timestamp and therefore enable the computation of preintegrated measurements for each lidar point.

Technically, given raw IMU readings $\acci$ and $\gyri$, GP regression \cite{Rasmussen2006} is used to infer $\acct$ and $\gyrt$ probabilistically at any time $\time$.
The use of a non-parametric method makes the UPMs independent of any explicit motion model.
Regressions are conducted independently for each IMU measurement DoF.
\def\imusample{s_j}
Let us denote $\imusample$ the $j^{th}$ DoF of the IMU readings to be interpolated. Using a GP model $\imusample\sim\mathcal{GP}(0,k(\time,\time'))$, the mean and variance of $\imusample$ are inferred as
\begin{align}
    \imusample^*(\time) &= \mathbf{k}(\time, \imutime)^\top ( \mathbf{K}(\imutime, \imutime) + \sigma^2_{\imusample} \mathbf{I})^{-1}\mathbf{\imusample}, \text{ and}
    \\
    {\sigma^2_{\imusample}}^*(\time) &= k(\time, \time) - \mathbf{k}(\time, \imutime)^\top ( \mathbf{K}(\imutime,\imutime) + \sigma^2_{\imusample} \mathbf{I})^{-1} \mathbf{k}(\time, \imutime),
\end{align}
with $\mathbf{\imusample}$ being the corresponding vector of training values for $\imusample$ (i.e. the raw intertial readings), $\sigma^2_{\imusample}$ the noise variance of the training data, $k$ the kernel covariance function, $\mathbf{k}(\time,\imutime)^\top = [\begin{smallmatrix} k(t, \imutime_1)&\cdots&k(t, \imutime_Q)\end{smallmatrix}]$, and $\mathbf{K}$ the matrix created by stacking the vectors $\mathbf{k}(\imutimei,\imutime)^\top$ with $i=1,\cdots,Q$.

To address the cubic complexity of the GP regression, only the samples in temporal windows aligned with the lidar frame scanning time (from $\frametimem$ to $\frametimemp$) are considered.
More rigorously, to infer the IMU readings for each of the points in $\scansetm$, training samples must have a timestamp $\imutimei \in [\frametimem - o; \frametimemp + o]$, with $o$ a time overlap over the adjacent frames.
In our implementation we use the Matern covariance function $k(t,t')\!=\!\sigma_k^2(1+\frac{\sqrt{3}\lVert t\text{-}t' \lVert}{l_k})\exp(-\frac{\sqrt{3}\lVert t\text{-}t' \lVert}{l_k})$.
The hyper-parameters $\sigma_{\imusample}$, $\sigma_k$, and $l_k$ are respectively initialised with the sensor variance from the manufacturer specification, the empirical variance of $\imusample$ over the window, and thirty times the IMU period.
Ultimately, these hyper-parameters are optimised for each of the training windows as described in \cite{Rasmussen2006}.

\section{Back-end}

\subsection{Lidar factors}

Lidar factors correspond to distance residuals computed between lidar feature-points and their corresponding feature-points from other lidar frames.
As we will explain in the front-end section, the set of feature associations $\association$ contains tuples of 3 (point-to-edge constraints) or 4 feature-points (point-to-plane constraints).

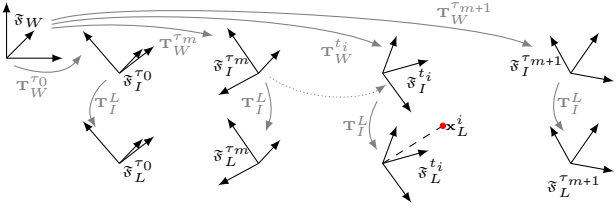
\begin{figure}
    \begin{tikzpicture}
    
    \def\framescale{0.75}

    \draw[->] (0,-0.3) -- (\framescale,-0.3);
    \draw[->] (0,-0.3) -- (0,-0.3+\framescale);
    \draw[->] (0,-0.3) -- (\framescale*0.5,-0.3+\framescale*0.5);
    \node at (0.3,0.2) {\tiny $\worldframe$};
    
    \draw[->] (1.5,-0.5) -- (1.5+\framescale*0.42,-0.5+\framescale*0.44);
    \draw[->] (1.5,-0.5) -- (1.5-\framescale*0.65,-0.5+\framescale*0.75);
    \draw[->] (1.5,-0.5) -- (1.5+\framescale*0.62,-0.5+\framescale*0.48);
        \node at (1.75,-0.6) {\tiny $\imuframe^{\frametime_0}$};
    
    \draw[->] (1.5,-1.7) -- (1.5+\framescale*0.42,-1.7+\framescale*0.44);
    \draw[->] (1.5,-1.7) -- (1.5-\framescale*0.65,-1.7+\framescale*0.75);
    \draw[->] (1.5,-1.7) -- (1.5+\framescale*0.62,-1.7+\framescale*0.48);
        \node at (1.75,-1.8) {\tiny $\lidarframe^{\frametime_0}$};
    
    \draw[->] (3.35,-0.5) -- (3.35-\framescale*0.56,-0.5+\framescale*0.81);
    \draw[->] (3.35,-0.5) -- (3.35+\framescale*0.39,-0.5+\framescale*0.42);
    \draw[->] (3.35,-0.5) -- (3.35-\framescale*0.72,-0.5-\framescale*0.39);
        \node at (3,-0.4) {\tiny $\imuframe^{\frametime_m}$};
    
    \draw[->] (3.35,-1.7) -- (3.35-\framescale*0.56,-1.7+\framescale*0.81);
    \draw[->] (3.35,-1.7) -- (3.35+\framescale*0.39,-1.7+\framescale*0.42);
    \draw[->] (3.35,-1.7) -- (3.35-\framescale*0.72,-1.7-\framescale*0.39);
        \node at (3,-1.6) {\tiny $\lidarframe^{\frametime_m}$};

    \draw[->] (5,-0.5) -- (5+\framescale*0.82,-0.5+\framescale*0.23);
    \draw[->] (5,-0.5) -- (5+\framescale*0.25,-0.5+\framescale*0.67);
    \draw[->] (5,-0.5) -- (5+\framescale*0.51,-0.5-\framescale*0.70);
    \node at (5.5,-0.6) {\tiny $\imuframe^{\timei}$};
    
    \draw[->] (5,-1.7) -- (5+\framescale*0.82,-1.7+\framescale*0.23);
    \draw[->] (5,-1.7) -- (5+\framescale*0.25,-1.7+\framescale*0.67);
    \draw[->] (5,-1.7) -- (5+\framescale*0.51,-1.7-\framescale*0.70);
    \node at (5.65,-1.8) {\tiny $\lidarframe^{\timei}$};
    
    \draw[->] (7.5,-0.5) -- (7.5+\framescale*0.50,-0.5+\framescale*0.72);
    \draw[->] (7.5,-0.5) -- (7.5-\framescale*0.37,-0.5+\framescale*0.67);
    \draw[->] (7.5,-0.5) -- (7.5+\framescale*0.77,-0.5-\framescale*0.14);
        \node at (7.05,-0.4) {\tiny $\imuframe^{\frametime_{m\text{+}1}}$};
    
    \draw[->] (7.5,-1.7) -- (7.5+\framescale*0.50,-1.7+\framescale*0.72);
    \draw[->] (7.5,-1.7) -- (7.5-\framescale*0.37,-1.7+\framescale*0.67);
    \draw[->] (7.5,-1.7) -- (7.5+\framescale*0.77,-1.7-\framescale*0.14);
        \node at (7.55,-2.0) {\tiny $\lidarframe^{\frametime_{m\text{+}1}}$};
    
    \draw[gray,->] (0.1,-0.4) .. controls (0.35,-0.5) and (0.75,-0.55) .. (1,-0.25);
        \node[gray] at (0.35,-0.68) {\tiny $\imuT^{\frametime_0}$};
    
    \draw[gray,->] (0.6,0.1) .. controls (1.5,0.2) and (2.25,0.1) .. (2.75,0);
        \node[gray] at (2.3,-0.12) {\tiny $\imuT^{\frametime_m}$};
    
    \draw[gray,->] (0.6,0.15) .. controls (1.5,0.35) and (4.0,0.25) .. (5,-0.15);
        \node[gray] at (4.4,-0.2) {\tiny $\imuT^{\timei}$};

    \draw[gray,->] (0.6,0.2) .. controls (1.5,0.5) and (6.0,0.25) .. (7,-0.15);
        \node[gray] at (6.1,0.3) {\tiny $\imuT^{\frametime_{m\text{+}1}}$};
    
    \node[gray] at (1.35,-0.9) {\tiny $\Tc$};
    \node (c00) at (1.45,-0.5) {};
    \node (c01) at (1.25,-1.35) {};
    \draw[->, gray] (c00) edge [bend right = 40] (c01);

    \node[gray] at (3.3,-0.9) {\tiny $\Tc$};
    \node (c10) at (3.4,-0.5) {};
    \node (c11) at (3.4,-1.4) {};
    \draw[->, gray] (c10) edge [bend left = 20] (c11);
   
    \node[gray] at (4.65,-1.2) {\tiny $\Tc$};
    \node (c20) at (5.0,-0.75) {};
    \node (c21) at (5.0,-1.65) {};
    \draw[->, gray] (c20) edge [bend right = 30] (c21);
    
    \node[gray] at (7.5,-0.9) {\tiny $\Tc$};
    \node (c30) at (7.4,-0.5) {};
    \node (c31) at (7.5,-1.4) {};
    \draw[->, gray] (c30) edge [bend right = 30] (c31);
    
    \draw[densely dotted,gray,->] (3.5,-0.55) .. controls (4,-0.90) and (4.75,-0.85) .. (5.05,-0.65);

    \draw[dashed] (5.0,-1.7) -- (5.8,-1.2);
    \filldraw[red] (5.8,-1.2) circle (1pt);
    \node at (6,-1.2) {\tiny $\lidarpointi$};
    
    \end{tikzpicture}
    \caption{Frames and frame transformations during a sequence of measurements. $\imuframem$ and $\lidarframem$ respectively represent the IMU and lidar frames at time $\frametime_m$. The grey continuous line arrows represent the transformations between the different frames. $\worldframe$ is the world fixed frame. The dotted line shows the use of upsampled preintegrated measurements to reproject the point $\lidarpointi$.}
    \label{figure:frames}
\end{figure}

For the lidar factors, point-to-line or point-to-plane distances are used.
The matched points found in $\association$ are projected in the world frame $\worldframe$ using the calibration parameters, UPMs for each of the points and the current estimates of the IMU poses and velocities (Fig.~\ref{figure:frames}).
Therefore, a point $\lidarpointi \in \scansetm$ is projected in $\worldframe$ using \eqref{eq:calib_projection}, \eqref{eq:imu_p} and \eqref{eq:imu_r},
\begin{equation}
    \begin{bmatrix}\worldpointi\\1\end{bmatrix} = \imuTi \Tc \begin{bmatrix}\lidarpointi\\1\end{bmatrix}.
\label{eq:to_world_frame}
\end{equation}

Let us denote an edge association $\oneassociation_3 \in \association$. $\oneassociation_3 = \{\lidarpointi, \lidarpointj, \lidarpointk\}$ with $\lidarpointi \in \feature^m$, $\lidarpointj \in \feature^n$, $\lidarpointk \in \feature^o$ and $n,o \neq m$. These points are projected in $\worldframe$ via \eqref{eq:to_world_frame} to get $\worldpointi$, $\worldpointj$ and $\worldpointk$.
The point-to-line distance
\begin{equation}
d_{\oneassociation_3} = \textstyle\frac{\lVert (\worldpointi - \worldpointj) \times (\worldpointi - \worldpointk) \lVert_2}{\lVert(\worldpointj - \worldpointk) \lVert_2}
\end{equation}
is used as an edge feature residual.

Let us denote a plane association $\oneassociation_4 \in \association$. $\oneassociation_4 = \{\lidarpointi, \lidarpointj, \lidarpointk, \lidarpointl\}$ with $\lidarpointi \in \feature^m$, $\lidarpointj \in \feature^n$, $\lidarpointk \in \feature^o$, $\lidarpointl \in \feature^p$ and $n,o,p \neq m$. These points are projected in $\worldframe$ via \eqref{eq:to_world_frame} to get $\worldpointi$, $\worldpointj$, $\worldpointk$ and $\worldpointl$.
The point-to-plane distance
\begin{equation}
d_{\oneassociation_4} = \textstyle\frac{ (\worldpointi - \worldpointj)^\top \big((\worldpointj - \worldpointk) \times (\worldpointj - \worldpointl)\big) }{\lVert(\worldpointj - \worldpointk) \times (\worldpointj - \worldpointl) \lVert_2}
\end{equation}
is used as a plane feature residual. As in \cite{LeGentil2018}, the variance of lidar residuals requires the knowledge of the state.
Therefore, the noise propagation $\boldsymbol{\Sigma}_{\lidarresiduala} = \mathbf{J}_{\lidarresiduala}(\state)\boldsymbol{\Sigma}_z(\mathbf{J}_{\lidarresiduala}(\state))^\top$, with $\boldsymbol{\Sigma}_z$ the covariance of the corresponding lidar and UPMs measurements, and $\mathbf{J}_{\lidarresiduala}(\state)$ the Jacobian of $\lidarresiduala$ with respect to the sensors measurements evaluated at the current best estimate of the state $\state$, needs to be executed regularly during the optimisation.

\subsection{IMU factors}

The IMU factors constitute direct constraints on the IMU poses and velocities. With $\Delta \frametimem = \frametimem - \frametimemm$, the associated residual $\imuresidualm = [ {\imurresidualm}; {\imuvresidualm}; {\imupresidualm}]$ is obtained directly by manipulating \eqref{eq:imu_p}, \eqref{eq:imu_v}, and \eqref{eq:imu_r},
\begin{align}
&\begin{aligned}
\imupresidualm\!=& {\imurotmm}^\top(\imuposm - \imuposmm - \Delta \frametimem \imuvelmm - \textstyle\frac{{\Delta \frametimem}^2}{2}\gravity) - \upmposmm
\nonumber
\end{aligned}
\\
&\begin{aligned}
\imuvresidualm\!=& {\imurotmm}^\top(\imuvelm - \imuvelmm - \Delta \frametimem\gravity) - \upmvelmm
\nonumber
\end{aligned}
\\
&\begin{aligned}
\imurresidualm\!=& \text{Log} ( {\upmrotmm}^\top {\imurotmm}^\top\imurotm).
\end{aligned}
\end{align}

\subsection{IMU biases and inter-sensor time-shift}

The UPMs computation \eqref{eq:imu_delta} is a function of the accelerometer biases $\bacc$, gyroscope biases $\bgyr$, and inter-sensor time-shift $\timeshift$.
However, these values are not perfectly known at the time of preintegration.
In our framework, we model the IMU biases as a Brownian motion as in \cite{Furgale2013} and the inter-sensor time-shift as a simple Gaussian.
By considering biases and time-shift locally constant during lidar frames, and by adopting a first-order expansion as in \cite{Forster2015a}, the UPMs can be approximated as:

\begin{align}
&\begin{aligned}
{\upmroti}{(\bgyr, \timeshift)} \approx& {\upmroti}{(\bgyrpriorm,\timeshiftpriorm)} \text{Exp}\Big(\textstyle\frac{\partial \upmroti}{\partial \bgyr} \bgyrcorrectionm\\
&+ \textstyle\frac{\partial \upmroti}{\partial \timeshift} \timeshiftcorrectionm \Big)
\nonumber
\\
{\upmveli}{(\bacc,\bgyr, \timeshift)} \approx& {\upmveli}{(\baccpriorm,\bgyrpriorm,\timeshiftpriorm)} + \textstyle\frac{\partial \upmveli}{\partial \bacc} \bacccorrectionm
\\ & + \textstyle\frac{\partial \upmveli}{\partial \bgyr} \bgyrcorrectionm  + \textstyle\frac{\partial \upmveli}{\partial \timeshift} \timeshiftcorrectionm
\end{aligned}
\\
&\begin{aligned}
{\upmposi}{(\bacc,\bgyr, \timeshift)} \approx& {\upmposi}{(\baccpriorm,\bgyrpriorm,\timeshiftpriorm)} + \textstyle\frac{\partial \upmposi}{\partial \bacc} \bacccorrectionm 
\\ &+ \textstyle\frac{\partial \upmposi}{\partial \bgyr} \bgyrcorrectionm  + \textstyle\frac{\partial \upmposi}{\partial \timeshift} \timeshiftcorrectionm,
\end{aligned}
\label{eq:biases_linearisation}
\end{align}
with $\baccm = \baccpriorm + \bacccorrectionm$, $\bgyrm = \bgyrpriorm + \bgyrcorrectionm$, and $\timeshiftm = \timeshiftpriorm + \timeshiftcorrectionm$.
Note that $\bar{\bullet}$ denotes the prior knowledge of the value at the time of preintegration and $\hat{\bullet}$ represents the correction.
In our implementation, the Jacobians of the UPMs with respect to the inter-sensor time-shift are the only Jacobians computed numerically as the derivative of the GP-interpolated inertial readings with respect to the time-shift are not readily available and kernel dependent.
The residuals 
\begin{align}
\baccresidualm = \baccpriorm + \bacccorrectionm - \baccpriormm - \bacccorrectionmm
\\
\bgyrresidualm = \bgyrpriorm + \bgyrcorrectionm - \bgyrpriormm - \bgyrcorrectionmm
\end{align}
are used in the biases factors to impose the Brownian motion constraint.
The time-shift factor residual is simply  $\timeshiftresidualm = \timeshiftcorrectionm$ as per the Gaussian noise model.

\section{Front-end}

The front-end of the proposed method aims at populating the set $\mathcal{A}$ of lidar feature-point associations to allow frame-to-frame and loop closure matching.

\subsection{Feature extraction}

\def\lidarline{\mathcal{N}}
\def\lidarlineml{\lidarline^m_l}
\def\leftset{\mathcal{L}}
\def\leftseti{\leftset_i}
\def\rightset{\mathcal{R}}
\def\rightseti{\rightset_i}
\def\azimuth{\alpha}
\def\localpointx{x_{P_i}}
\def\localpointy{y_{P_i}}
\def\lastlidarpointik{\lastlidarpoint^{i+k}}
\def\regressionx{\mathbf{X}}
\def\regressionxleft{\regressionx_{\leftseti}}
\def\regressionxright{\regressionx_{\rightseti}}
\def\regressiony{\mathbf{Y}}
\def\regressionyleft{\regressiony_{\leftseti}}
\def\regressionyright{\regressiony_{\rightseti}}
\def\regressionerrori{\bar{e}^i}
\def\maxregressionerrori{e^i}
\def\regressionerrorrighti{\regressionerrori_{\rightset}}
\def\maxregressionerrorrighti{\maxregressionerrori_{\rightset}}
\def\regressionerrorlefti{\regressionerrori_{\leftset}}
\def\maxregressionerrorlefti{\maxregressionerrori_{\leftset}}
\def\thresholdregressionerror{\bar{e}_{th}}
\def\thresholdmaxregressionerror{e_{th}}
\def\planefeature{\mathcal{P}}
\def\planefeaturem{\planefeature^m}
\def\planefeaturei{\planefeature^i}
\def\edgeinfeature{\mathcal{E}_I}
\def\edgeoutfeature{\mathcal{E}_O}
\def\edgeinfeaturem{\edgeinfeature^m}
\def\edgeoutfeaturem{\edgeoutfeature^m}
\def\edgeinfeaturei{\edgeinfeature^i}
\def\edgeoutfeaturei{\edgeoutfeature^i}
\def\regressionslope{s}
\def\regressionintercept{q}
\def\directionvector{\mathbf{v}}
\def\scorei{c_i}

This subsection of the proposed method has been described in the conference paper \cite{LeGentil2019}, but for completeness, it is also described here with additional details.

The vertical resolution of most of today's lidars has driven the design of our feature extraction algorithm toward a channel-by-channel method in a similar way to the one in \cite{Zhang2014b}.
The authors of \cite{Zhang2014b} introduced a computationally efficient smoothness score for feature extraction/classification.
While robust in weakly structured environments and allowing for real-time operations, this score computation is not fully consistent.
For example, different points belonging to the same planar surface will have different smoothness scores despite the same underlying structure.
We propose a feature extraction technique based on linear regression to describe the surface observed by the lidar consistently.

Given an N-channel lidar, each lidar scan $\scansetm$ is split into $N$ ``lines", $\lidarlineml$ ($l = 1,\cdots,N$), according to the elevation of the 3D-points collected. All the points are given a curvature score.
The curvature computation aims at fitting lines to two subsets of points adjacent to the point under examination $\lidarpointi \in \lidarlineml$, and then to retrieve the cosine of the angle between these two lines.
The subsets, $\leftseti$ and $\rightseti$ contain the $D$ previous and following measurements (with respect to $\lidarpointi$) in $\lidarlineml$.

First, the points need to be reprojected into the lidar frame at $\frametimem$ ($\lidarframem$) to remove motion distortion according to the best current estimate of the state $\state$.
These reprojected points $\lastlidarpointi$ are computed as follows:
\begin{equation}
    \begin{bmatrix}\lastlidarpointi\\1\end{bmatrix} = (\Tc)^{-1} (\imuTm)^{-1} \imuTi\Tc\begin{bmatrix}\lidarpointi\\1\end{bmatrix}.
\label{eq:to_lidar_frame}
\end{equation}

\begin{figure}
    \centering
    \newcommand\parta[2]{
            \node[anchor=south west,inner sep=0] at (#1+0,#2+0) {\includegraphics[clip, trim=0cm 0.28cm 0cm 0.28cm, width=0.33\columnwidth,]{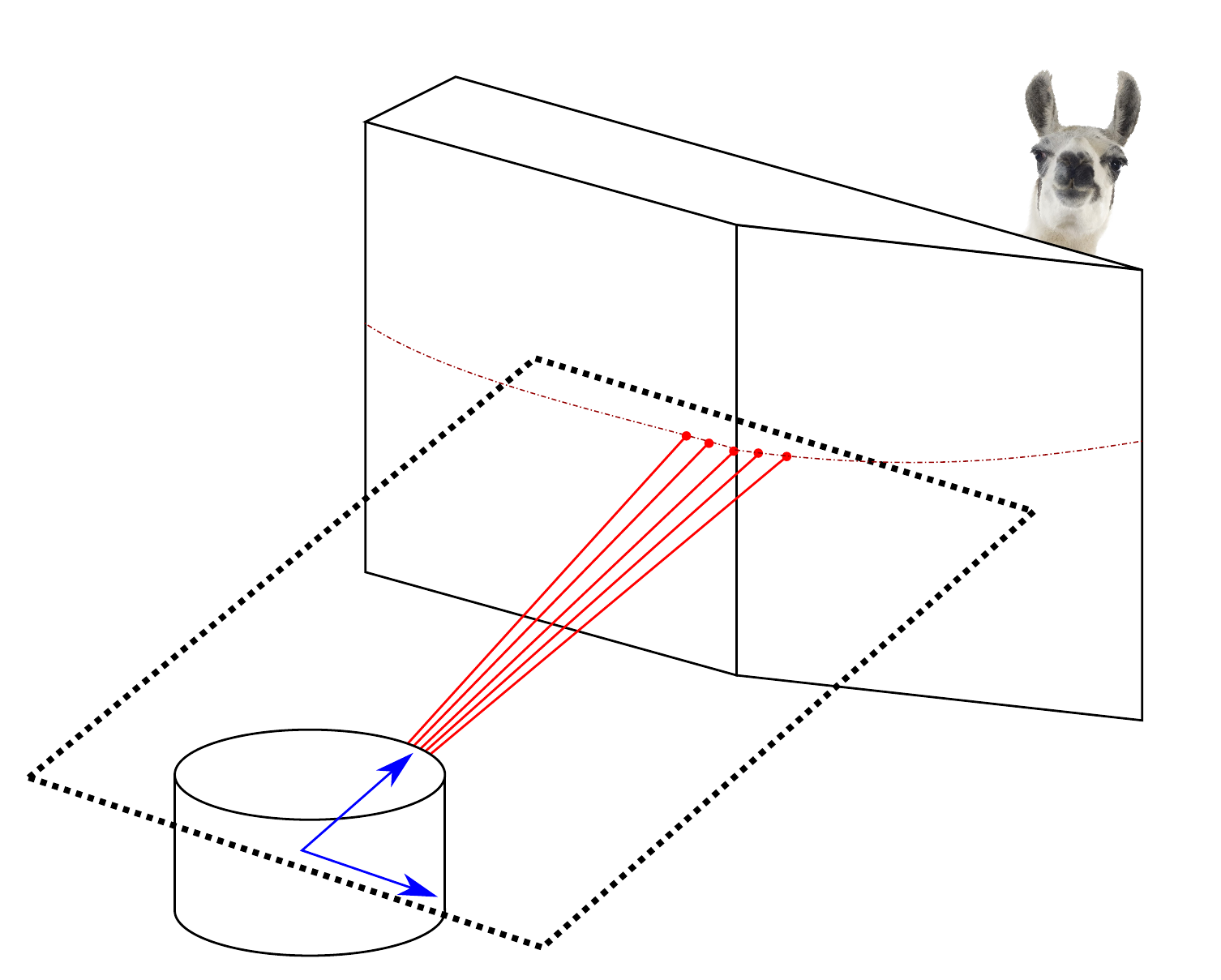}};

            \node at (#1+0.2,#2+0.95) [black!80]{\scriptsize $Lidar$};
            \node at (#1+1.8,#2+2.05) [gray, rotate=-16]{\scriptsize $Wall$};
            \node at (#1+0.25,#2+1.60) [gray, align=left]{\scriptsize $Local$};
            \node at (#1+0.25,#2+1.40) [gray, align=left]{\scriptsize $plane$};

            \node at (#1+1.3,#2+0.3) [blue]{\scriptsize $\localpointx$};
            \node at (#1+0.85,#2+0.65) [blue]{\scriptsize $\localpointy$};

            \node at (#1+0.6,#2+2) [red]{\scriptsize $\lidarlineml$};
            \node (Dn) at (#1+1.12, #2+1.35) {};
            \node (En) at (#1+0.6, #2+1.9) {};
            \draw[red,->] (En) edge (Dn);

            \node (D) at (#1+0.9, #2+0.9) {};
            \node (E) at (#1+0.4, #2+1.4) {};
            \draw[gray,->] (E) edge (D);
            \node (F) at (#1+0.60, #2+0.37) {};
            \node (G) at (#1+0.1, #2+0.9) {};
            \draw[black!80,->] (G) edge (F);
    }
        \newcommand\partb[2]{
            \node[anchor=south west,inner sep=0] at (#1+0,#2+0) {\includegraphics[clip,trim=0cm 0.28cm 0cm 0.28cm, width=0.33\columnwidth]{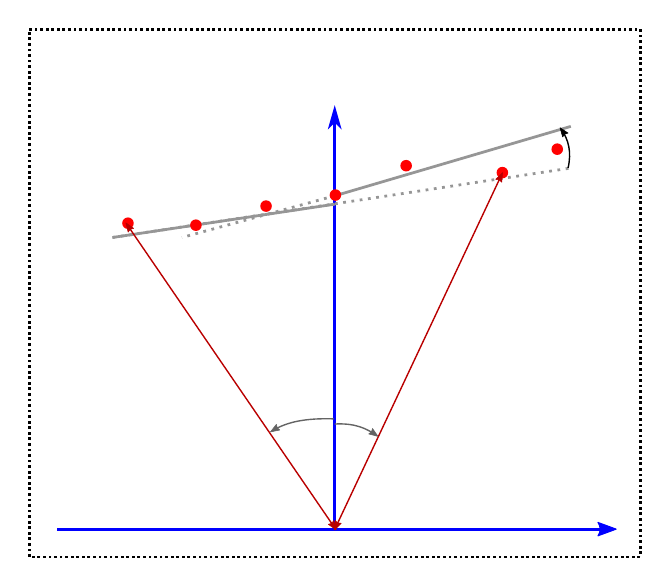}};

            \node at (#1+0.9,#2+2.15) [gray]{\scriptsize $Local\ plane$};
            \node at (#1+1.05,#2+1.0) [gray, rotate=-55]{\scriptsize $\azimuth^{i\text{-}3}\text{-}\azimuth^i$};
            \node at (#1+1.75,#2+1.1) [gray, rotate=65]{\scriptsize $\azimuth^{i\text{+}2}\text{-}\azimuth^i$};

            \node at (#1+0.7,#2+0.6) [red]{\scriptsize $\lvert \lastlidarpoint^{i\text{-}3}\lvert$};
            \node at (#1+2.2,#2+0.6) [red]{\scriptsize $\lvert \lastlidarpoint^{i\text{+}2}\lvert$};

            \node at (#1+2.6,#2+0.25) [blue]{\scriptsize $\localpointx$};
            \node at (#1+1.75,#2+1.9) [blue]{\scriptsize $\localpointy$};

            \node at (#1+2.65,#2+1.8) [black]{\scriptsize $\beta$};
        }
        \newcommand\partc[2]{
            \node[] (p1) at (#1+1.0,#2+1){};
            \node[] (p3) at (#1+1.0,#2+0){};
            \node[circle,fill,green,draw,inner sep=0pt,minimum size=3pt] (p2) at (#1+1.3,#2+0.5){};

            \draw[black] (#1+0,#2+0.5) circle (0.15cm);
            \draw (p1) -- (p2) -- (p3);
            \node at (#1+0,#2+0.8) [black!80]{\scriptsize $Lidar$};
            \node at (#1+1,#2+0.9) [anchor=west,gray]{\scriptsize $Wall$};
            \node at (#1+0.6,#2-0) [greentam]{\scriptsize Outward edge};
        }
        \newcommand\partd[2]{
            \node[] (p1) at (#1+1.3,#2+1){};
            \node[] (p3) at (#1+1.3,#2+0){};
            \node[circle,fill,blue,draw,inner sep=0pt,minimum size=3pt] (p2) at (#1+1.0,#2+0.5){};

            \draw[black] (#1+0,#2+0.5) circle (0.15cm);
            \draw (p1) -- (p2) -- (p3);
            \node at (#1+0.63,#2) [bluetam]{\scriptsize Inward edge};
        }
        \begin{tikzpicture}
            \parta{0}{0};
            \partb{3.1}{0};
            \partc{6.7}{1.2};
            \partd{6.7}{0.1};
            \node at (1.5,-0.2) [black]{\small(a)};
            \node at (4.55,-0.2) [black]{\small(b)};
            \node at (7.35,-0.2) [black]{\small(c)};
        \end{tikzpicture}
    \caption{
        Geometric feature extraction based on linear regression. (a):~The points around a given azimuth are assumed to belong to a local plane. (b):~ On that local plane, linear regressions are performed considering points in $\lidarlineml$ on both sides of the $i^{th}$ point $\lidarpointi \in\lidarlineml$ independently. The curvature score is equal to $\cos(\beta)$ with $\beta$ the angle between the two fitted lines. (c):~Edge classification for data association robustness.}
    \label{figure:feature_extraction}
\end{figure}
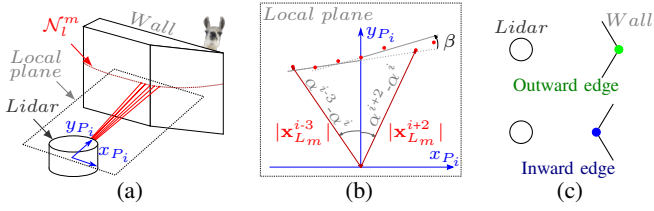

The curvature scores are computed under the approximation that around a certain azimuth the consecutively measured 3D-points belong to the same plane.
As shown in Fig.~\ref{figure:feature_extraction}, and given $\azimuth^i$ the new azimuth of $\lastlidarpointi$, the points in $\leftseti$ and $\rightseti$ are projected on a plane space around $\azimuth^i$.
The points' coordinates in the plane, $\localpointx^k$ and $\localpointy^k$, are computed as
\begin{align}
    \begin{bmatrix} 
        \localpointx^k,\,\localpointy^k
    \end{bmatrix}\! =\! \lvert \lastlidarpointik\lvert \begin{bmatrix} \sin (\azimuth^{i+k} - \azimuth^i),\,\cos (\azimuth^{i+k} - \azimuth^i)\end{bmatrix},
\label{eq:local_plane}
\end{align}
with $k = -D,\cdots, D$ ($D = 5$ in our implementation).

\begin{align}
	\regressionxleft\! &= \!\begin{bmatrix}
        \localpointx^{-D} & \cdots & \localpointx^{0} \\
        1 & \cdots & 1
	\end{bmatrix}^\top\!\!,
	\regressionxright\! = \!\begin{bmatrix}
        \localpointx^{0} & \cdots & \localpointx^{D} \\
        1 & \cdots & 1
	\end{bmatrix}^\top\!\!,
	\\
	\nonumber
	\regressionyleft\! &= \!\begin{bmatrix}
	\localpointy^{-D} &
	\cdots &
	\localpointy^{0}
	\end{bmatrix}^\top
	\text{and }
	\regressionyright\! =\! \begin{bmatrix}
	\localpointy^{0} &
	\cdots &
	\localpointy^{D}
	\end{bmatrix}^\top
\end{align}
group the projected points coordinates according to the two adjacent subsets $\leftseti$ and $\rightseti$. In the rest of this section, $\bullet$ represents either $\leftseti$ or $\rightseti$.
A line of slope $s_\bullet$ and y-intercept $q_\bullet$ can be fitted to the subset $\bullet$ with 
\begin{equation}
\begin{bmatrix}
\regressionintercept_{\bullet} &
\regressionslope_{\bullet}
\end{bmatrix}^\top
= \big(\regressionx_\bullet^\top \regressionx_\bullet \big)^{-1} \regressionx_\bullet^\top \regressiony_\bullet,
\end{equation}
and an associated unit direction vector can be obtained as
\begin{equation}
\directionvector_{\bullet} = \begin{bmatrix}
\frac{1}{\sqrt{1+\regressionslope_{\bullet}^2}} &
\frac{\regressionslope_{\bullet} }{ \sqrt{1+\regressionslope_{\bullet}^2}}
\end{bmatrix}^\top.
\end{equation}
The average and maximum regression error values
\begin{align}
    \regressionerrori_{\bullet} &= \frac{1}{\lvert \bullet \lvert} \sum_{k|\lidarpointk\in \bullet} \Big| \localpointy^k - \regressionintercept_{\bullet} - \regressionslope_{\bullet}\localpointx^k\Big| \text{ and}\\
    \maxregressionerrori_{\bullet} &= \max_{k|\lidarpointk \in \bullet} \big( \big| \localpointy^k - \regressionintercept_{\bullet} - \regressionslope_{\bullet}\localpointx^k\big|\big)
\end{align}
are used to reject points or to detect border of occlusions as per the algorithm described in the Appendix.
The score $\scorei = \directionvector_{\leftseti}^\top \directionvector_{\rightseti}^{}$  represents the cosine of the angle between the two fitted lines. Thus, $\scorei$ is close to $1$ when the underlying surface is planar, and decreases with the sharpness of edges.

As in \cite{Zhang2014b}, surfaces close to being parallel to the laser beams are rejected as features.
We also use a system of bins and a maximum number of features per bin on each laser line to ensure the features  are spread over the whole scan.
The points with the highest scores in each of the bins of $\lidarlineml$ are classified as planar points and the lowest scores as edges according to arbitrarily chosen maximum numbers of features per bin and thresholds on scores.
The edge orientation (Fig.~\ref{figure:feature_extraction}\,(c)), classified as inward (pointing toward the lidar) or outward (pointing away from the lidar), can be defined by looking at the values of the regressed lines' parameters.
This edge classification brings additional robustness to the later feature association as inward and outward edges cannot be matched together.
All the planar features in  $\lidarlineml$ with $l = 1,\cdots,N$, are grouped into a set $\planefeaturem$, the inward edges in $\edgeinfeaturem$ and outward edges in $\edgeoutfeaturem$.
The reader should note that the feature set (from the back-end section of this paper) $\featurem = \planefeaturem \cup \edgeinfeaturem \cup \edgeoutfeaturem$.

\subsection{Feature recomputation}

\def\nfeatures{N_f}
\def\checksetj{\mathcal{C}^j_{m,k}}
\def\checksetjm{\mathcal{C}^{j-1}_{m,k}}
\def\binset{\mathcal{B}_{m,k}}
\def\binsetj{\binset^j}
\def\binsetjm{\binset^{j-1}}

The aforementioned process of feature extraction is computationally costly and depends on the last estimate of the state $\state$. IN2LAAMA integrates a way to check the validity of features without the need to recompute all the linear regressions.

For the moment, let us consider planar features only and define $\nfeatures$ as the maximum number of planar features selected per bin during the feature extraction performed on the $\mth$ lidar frame.
The set of planar features in the $k^{th}$ bin of the $\mth$ lidar frame is denoted $\binsetj$ with $j>0$ corresponding to the $j^{th}$ time the features of frame $m$ have been computed.
Considering the case $j=1$, the scores $\scorei$ are computed for all points in $\scansetm$. The points are then sorted according to their score in a decreasing order.
Starting from the highest score, points are added to $\binsetj$ if their score is above a threshold and as long as $\lvert \binsetj \lvert < \nfeatures$.
The algorithm also stores the $\nfeatures$ next candidates (even if they do not match the threshold) in the set $\checksetj$.
Note that the set union of the planar feature bins $\binsetj$ is $\planefeaturem$.

In the case of $j>1$, typically after an optimisation iteration of the factor graph, the state $\state$ changes. The features potentially need to be recomputed.
The scores of the points in $\binsetjm$ and $\checksetjm$ are recomputed and sorted by decreasing order.
The point selection for the bins is done as if $j=1$, but using point scores from $\binsetjm \cup \checksetjm$ and not from $\scansetm$.
An overlap ratio of number of features is computed as
\begin{equation}
    \Theta_{j,m} = \textstyle\frac{\lvert (\bigcup\limits_{k}\binsetj) \cap (\bigcup\limits_{k}\binsetjm ) \lvert}{\lvert ( \bigcup\limits_{k}\binsetjm) \lvert}.
\end{equation}
If $\Theta_{j,m}$ is close to one, then $\binsetj \gets \binsetjm$ and $\checksetj \gets \checksetjm$. Otherwise, $\binsetj$ and $\checksetj$ are recomputed from $\scansetm$ as per the case $j=1$.
Similar process is used for edge features.

\subsection{Data association}

The proposed scan registration method requires matching features from frame-to-frame.
Feature matching is usually prone to outliers.
Thus, a robust process for data association is needed.
This section describes the different processes used in IN2LAAMA for matching and outlier rejection.

\subsubsection{Feature matching}
Given a pair of lidar frames $i$ and $m$ reprojected into $\worldframe$, the method looks for the 3 nearest neighbours of each point from $\planefeaturei$ in $\planefeaturem$.
For points in $\edgeinfeaturei$ and $\edgeoutfeaturei$, only the 2 nearest neighbours are searched in $\edgeinfeaturem$ and $\edgeoutfeaturem$ respectively. 
In both cases, to limit the impact of the measurements' noise on the point-to-line and point-to-plane distances used as lidar residuals, the $n = \{2,3\}$ closest points need to be spatially spread over some minimum distances.
The $n$ closest points cannot belong to a single lidar channel. If the $n$ closest points do not satisfy these conditions, the subsequent closest points are considered.
For planar feature associations, the collinearity of the 3 points from $\planefeaturem$ is checked.
Kd-trees \cite{Bentley1975} are used for efficient nearest neighbour searches.
The data associations are included in $\association$ as tuples of 3 or 4 as per the type of feature.

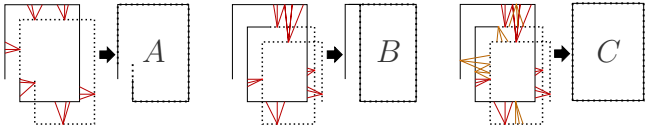
\begin{figure}
    \centering
        \def\letterx{2.1}
        \def\lettery{1.0}
    \begin{tabular}[t]{l c r}
        \begin{tikzpicture}
            \node[anchor=south west,inner sep=0] at (0,0) {\includegraphics[clip, trim=0cm 0cm 0cm 0cm, width=0.28\columnwidth]{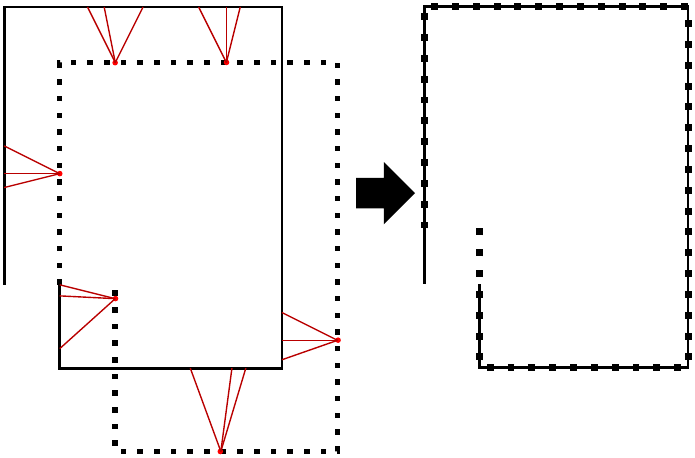}};
            \node at (\letterx-0.1,\lettery) [black!80]{\large $A$};
            \end{tikzpicture}
            \hfill
            &\begin{tikzpicture}
            \node[anchor=south west,inner sep=0] at (0,0) {\includegraphics[clip, trim=0cm 0cm 0cm 0cm, width=0.28\columnwidth]{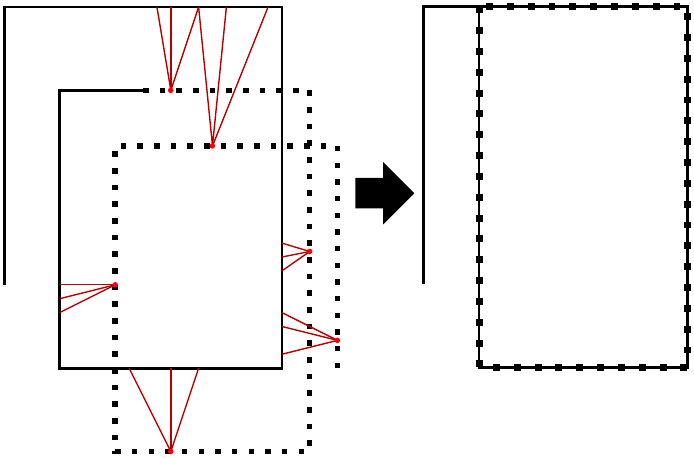}};

            \node at (\letterx,\lettery) [black!80]{\large $B$};
            \end{tikzpicture}
            \hfill
            &\begin{tikzpicture}
            \node[anchor=south west,inner sep=0] at (0,0) {\includegraphics[clip, trim=0cm 0cm 0cm 0cm, width=0.28\columnwidth]{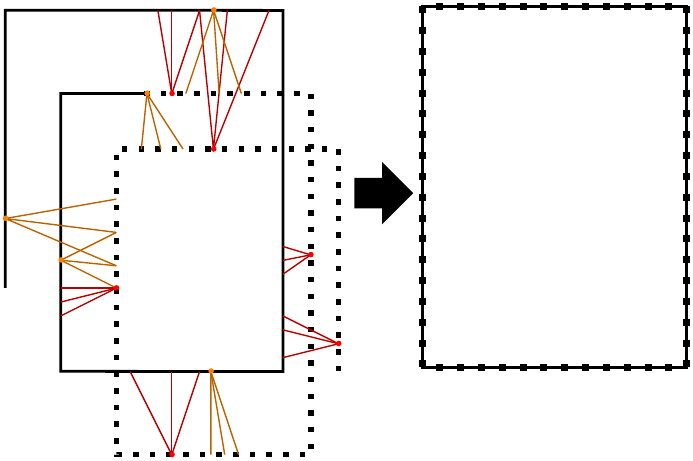}};

            \node at (\letterx-0.1,\lettery) [black!80]{\large $C$};
        \end{tikzpicture}
    \end{tabular}
    \caption{
    Different data association strategies between consecutive scans $\scansetmm$ (plain line) and $\scansetm$ (dashed line). For each scenario: left represents the data association; right shows the potential results after minimising point-to-plane distances. $A$ uses 360${}^\circ$ scans with back-association. $B$ uses scans greater than 360${}^\circ$, with back-association. $C$ extends $B$ with back-and-forth-association. $C$ ensure consistency of the lidar scans, whereas $A$ and $B$ do not.}
    \label{figure:frame_consistency}
\end{figure}

To enforce lidar scans' consistency without additional constraints, the proposed method considers scans greater than 360${}^\circ$ (520${}^\circ$ in our implementation) and does the data association both from $i$ to $m$, and from $m$ to $i$.
The idea behind these choices is illustrated in Fig.~\ref{figure:frame_consistency} through a 2D example in which the system moves in a rectangular room detecting only planar features and leveraging only lidar factors.
In scenario A, the data association between scans of 360${}^\circ$ (or less) does not necessarily allow for the correction of the motion distortion in the scans.
Individually the lidar residuals do not robustly constrain the motion distortion present in each frame, and the sensor noise can worsen this situation even more.
Without constraints on pose continuity between the last point of $\scansetmm$ and the first one of $\scansetm$, the registration is unlikely to correct the distorted scans properly.
In scenario B, the scans are greater than 360${}^\circ$, but the lower wall of the first scan does not appear in any data association.
Consequently, under some particular circumstances, the registration can still contain some motion distortion in $\scansetmm$.
The scenario C fully constrains the scan consistency, correcting the motion distortion by having both scans greater than 360${}^\circ$ combined with back-and-forth data association.

Intuitively, the greater the angle swept by a scan, the better the scan consistency.
Nonetheless, there is a trade-off with the execution time as the UPMs' GP regressions are computed from the IMU readings that are collected during the lidar scan.
Therefore, larger lidar scans imply cubically longer inference time as per the $\mathcal{O}(n^3)$ complexity of GP interpolation.

\subsubsection{Outliers rejection}
\def\planecheckset{\mathcal{U}}

To remove outliers from $\association$, matching points spread over too large areas are disregarded.
For planar features, the outlier detection additionally analyses the patch around the matched feature-points.
Considering a planar point $\worldpointi$ associated with $\worldpointj$, $\worldpointk$, and $\worldpointl$, the line-neighbours of each of the 3 matched points from $\planefeaturem$ ($\leftset_j$, $\rightset_j$, $\leftset_k$, $\rightset_k$, $\leftset_l$, and $\rightset_l$) are placed in a set $\planecheckset$.
If the points in $\planecheckset$ do not belong to the plane described by $\worldpointj$, $\worldpointk$, and $\worldpointl$, the association is rejected.
Formally, an association is valid if 
\begin{equation}
    \max_{\worldpoint^u \in \planecheckset} \Big( \textstyle\frac{ (\worldpoint^u - \worldpointj)^\top \big((\worldpointj - \worldpointk) \times (\worldpointj - \worldpointl)\big) }{\lVert(\worldpointj - \worldpointk) \times (\worldpointj - \worldpointl) \lVert_2} \Big)
\end{equation}
complies with the lidar range noise.

For edge features, the following two nearest neighbours in $\edgeinfeaturem$ or $\edgeoutfeaturem$ (depending on the edge orientation) are queried.
If not all the four neighbours lie on the same line (with compliance to the lidar range noise), the feature association is rejected.

\subsection{Loop-closure detection}

Loop-closures allow localisation and mapping algorithms to correct the accumulated drift inherent to frame-to-frame trajectory estimation.
The proposed method does not address large drift scenarios (such as the kidnapped robot scenario). It is out of the scope of this article and part of the future work.
Here, a simple geometric loop-closure detection based on estimated poses proximity has been implemented.
In other words, given the estimated state, if two poses are spatially close enough, a lidar factor is built between these two poses (performing feature matching and adding a set of residuals to the cost functions).
As an intuition, for indoor scenarios, it aims at detecting loop closures when the drift is smaller than the dimensions of the rooms.
An optional ICP test is conducted to validate or reject a loop closure candidate.
In the proposed method, loop-closures are modelled with additional lidar factors, as shown in Fig.~\ref{figure:factor_graph}.

\begin{figure}
    \centering
    \def\arrowlength{0.4}
    \newcommand\posea[2]{
        \coordinate (s0) at (0.75+#1,0.55+#2);
        \coordinate (s1) at (1.5+#1,0.85+#2);
        \coordinate (s2) at (1.5+#1,0.25+#2);
        \coordinate (s3) at (0.0+#1,0.85+#2);
        \coordinate (s4) at (0.0+#1,0.25+#2);
        \draw[fill=red!10,dashed] (#1,0.25+#2) rectangle (1.5+#1,0.85+#2);
        \filldraw[fill=red!20] (s0) -- (s1) -- (s2) -- cycle;
        \filldraw[fill=red!20] (s0) -- (s3) -- (s4) -- cycle;
        \draw[green,->] (0.75+#1,0.55+#2) -- (0.75+\arrowlength+#1,0.55+#2);
        \draw[blue,->] (0.75+#1,0.55+#2) -- (0.75+#1,0.55+\arrowlength+#2);
        \node at (0.75+#1,0.55+#2)[white,circle,fill,inner sep=1.25pt]{};
        \node at (0.75+#1,0.55+#2)[red,circle,fill,inner sep=0.5pt]{};
        \node at (0.75+#1,0.55+#2)[draw,red,circle,inner sep=1.25pt]{};
        \draw (#1,#2) rectangle (1.5+#1,1.1+#2);
    }
    \newcommand\poseb[2]{
        \coordinate (s0) at (0.75+#1,0.55+#2);
        \coordinate (s1) at (1.5+#1,0.85+#2);
        \coordinate (s2) at (1.5+#1,0.25+#2);
        \coordinate (s3) at (0.0+#1,0.85+#2);
        \coordinate (s4) at (0.0+#1,0.25+#2);
        \draw[fill=red!10,dashed] (#1,0.25+#2) rectangle (1.5+#1,0.85+#2);
        \filldraw[fill=red!20] (s0) -- (s1) -- (s2) -- cycle;
        \filldraw[fill=red!20] (s0) -- (s3) -- (s4) -- cycle;
        \draw[blue,->] (0.75+#1,0.55+#2) -- (0.75+#1,0.55+\arrowlength+#2);
        \draw[red,->] (0.75+#1,0.55+#2) -- (0.75-\arrowlength+#1,0.55+#2);
        \node at (0.75+#1,0.55+#2)[white,circle,fill,inner sep=1.25pt]{};
        \node at (0.75+#1,0.55+#2)[green,circle,fill,inner sep=0.5pt]{};
        \node at (0.75+#1,0.55+#2)[draw,green,circle,inner sep=1.25pt]{};
        \draw (#1,#2) rectangle (1.5+#1,1.1+#2);
    }
    \newcommand\posec[2]{
        \coordinate (s0) at (0.75+#1,0.55+#2);
        \coordinate (s1) at (0.97+#1,0.0+#2);
        \coordinate (s2) at (0.53+#1,0.0+#2);
        \coordinate (s3) at (0.97+#1,1.1+#2);
        \coordinate (s4) at (0.53+#1,1.1+#2);
        \draw[fill=red!10,dashed] (0.53+#1,0+#2) rectangle (0.97+#1,1.1+#2);
        \filldraw[fill=red!20] (s0) -- (s1) -- (s2) -- cycle;
        \filldraw[fill=red!20] (s0) -- (s3) -- (s4) -- cycle;
        \draw[green,->] (0.75+#1,0.55+#2) -- (0.75+#1,0.55+\arrowlength+#2);
        \draw[blue,->] (0.75+#1,0.55+#2) -- (0.75-\arrowlength+#1,0.55+#2);
        \node at (0.75+#1,0.55+#2)[white,circle,fill,inner sep=1.25pt]{};
        \node at (0.75+#1,0.55+#2)[red,circle,fill,inner sep=0.5pt]{};
        \node at (0.75+#1,0.55+#2)[draw,red,circle,inner sep=1.25pt]{};
        \draw (#1,#2) rectangle (1.5+#1,1.1+#2);
    }
    \newcommand\legend[2]{
        \draw[fill=red!10] (0.0+#1,0.0+#2) rectangle (0.5+#1,0.2+#2);
        \node at (0.5+#1,0.1+#2)[anchor=west]{\footnotesize Lidar FoV};
        \draw[blue,->] (2.3+#1,0+#2) -- (2.3+#1,0+0.75*\arrowlength+#2);
        \draw[green,->] (2.3+#1,0+#2) -- (2.3+0.75*\arrowlength+#1,0+#2);
        \node at (2.3+#1,0.0+#2)[white,circle,fill,inner sep=1.25pt]{};
        \node at (2.3+#1,0.0+#2)[red,circle,fill,inner sep=0.5pt]{};
        \node at (2.3+#1,0.0+#2)[draw,red,circle,inner sep=1.25pt]{};
        \node at (2.15+#1,0.05+#2){\color{red}\footnotesize$x$};
        \node at (2.60+#1,0.15+#2){\color{green}\footnotesize$y$};
        \node at (2.42+#1,0.25+#2){\color{blue}\footnotesize$z$};
        \node at (2.7+#1,0.1+#2)[anchor=west]{\footnotesize Lidar frame $\lidarframe$};
    }
    \begin{tikzpicture}
        \posea{0}{0.8}
        \poseb{2}{0.8}
        \posea{4.5}{0.8}
        \posec{6.5}{0.8}
        \legend{1.8}{0.1}
        \draw (3.95,0.6) -- (3.95,1.85);
        \draw (4.05,0.6) -- (4.05,1.85);
        \node at (5.25,1.82) [black!65,anchor = south]{\footnotesize Scan A};
        \node at (7.25,1.82) [black!65,anchor = south]{\footnotesize Scan B};
        \node at (0.75,1.82) [black!65,anchor = south]{\footnotesize Scan A};
        \node at (2.75,1.82) [black!65,anchor = south]{\footnotesize Scan B};
        \node at (1.75,0.33) [black,anchor = south]{\footnotesize $(a)$ $90^\circ$ rotation around $z$};
        \node at (6.25,0.33) [black,anchor = south]{\footnotesize $(b)$ $90^\circ$ rotation around $x$};
    \end{tikzpicture}
    \caption{Illustration of two loop-closure scenarios with different orientation gaps between the lidar scans. The sensing system is moved inside a room represented by the outer rectangle. In the scenario (a), there is a lot of geometric overlap between scan A and B. Despite the same amount of rotation, the scans registration in (b) is much more challenging because of the risk of poor data association due the small overlap.}
    \label{figure:loop_closure_angle}
\end{figure}
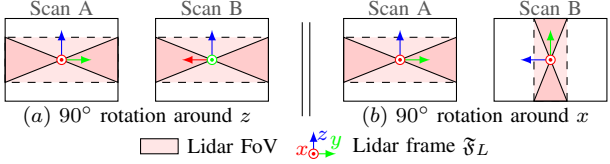

Commonly used lidars have a 360$^\circ$ field-of-view (FoV) around the spinning axis (azimuth) but have a narrower angular range on the other axis (elevation).
As illustrated in Fig.~\ref{figure:loop_closure_angle}, the nature of that set-up results in big overlaps between scans that have been collected while the lidar rotates around its spinning axis.
But if the lidar rotates around other axes, the overlap decreases, making the scan registration more challenging.
Therefore, the direct angle between two orientations cannot be used as part of the proximity metric. The 360$^\circ$ ``horizontal" FoV of the lidar must be taken into account.

\def\radialdist{d_r}
\def\radialcoordinate{d_h}
\def\axisdist{d_\alpha}
\def\xoriginframei{{x_m^i}}
\def\yoriginframei{{y_m^i}}
\def\zoriginframei{{z_m^i}}
\def\unitz{{\mathbf{u}_z}}

Let us consider a spinning lidar that sweeps the environment around the z-axis of its reference frame. The origin of the frame coincides with the lidar optical centre.
The different metrics used to define the closeness between two lidar frames $\lidarframem$ and $\lidarframei$ are as follows:
\begin{itemize}
	\item $\radialdist$ is the radial distance of the origin of $\lidarframei$ regarding the z-axis of $\lidarframem$.
	\item $\radialcoordinate$ is the point-to-plane distance between the origin of $\lidarframei$ and the plane formed by the x and y axes of $\lidarframem$.
	\item $\axisdist$ is the angle between the z-axes $\lidarframei$ and $\lidarframem$ when their origins coincide.
\end{itemize}

More formally, defining  $\unitz = [\begin{smallmatrix} 0&0&1 \end{smallmatrix}]^\top$ and $[\begin{smallmatrix}\xoriginframei&\yoriginframei&\zoriginframei & 1 \end{smallmatrix}]^\top = (\Tc)^{-1}(\imuTm)^{-1}\imuTTi [\begin{smallmatrix}\pc\\ 1 \end{smallmatrix}]$:
\begin{align}
    \radialdist& = \sqrt{{\xoriginframei}^2 + {\yoriginframei}^2},\qquad	\radialcoordinate = \lvert \zoriginframei \lvert\nonumber\\
    \text{and }&\cos(\axisdist) = {\unitz}^\top{\rc}^\top{\imurotm}^\top\imurotTi\rc\unitz.
\end{align}

To limit the number of redundant loop-closures, IN2LAAMA sets a minimum time between consecutive loop-closures, as well as a minimum gap time between the two frames used for a closure.
The algorithm looks for loop-closures every time a new frame is added to the factor graph.
The metrics $\radialdist$, $\radialcoordinate$, and $\lvert\cos(\axisdist)\lvert$ are computed between this new frame and the previous frames, that satisfy the aforementioned time conditions, by order of increasing timestamp.
The first frame that complies with thresholds on the above metrics is considered as a valid loop-closure candidate.
The threshold on $\radialdist$ is dynamically computed for each frame and is equal to the upper 1-sigma bound (mean plus standard deviation) of the range measurements of that frame.
The threshold on $\radialcoordinate$ is user-defined according to the environment and trajectory type, and the one on $\lvert\cos(\axisdist)\lvert$ is set according to the lidar vertical FoV ($2/3$ of the vertical FoV in our implementation).
Optionally, at the expense of longer computations, a standard ICP \cite{Segal2009} is conducted between the new frame and frames contained in a time window around the loop-closure candidate.
In this case, the loop-closure is validated only if the ICP fitness score is below a given value.
A valid loop-closure leads to the addition of a new lidar factor in the factor graph as it can be seen in Fig.~\ref{figure:factor_graph}.

\section{On the factor graph and implementation}

\subsection{Localisation and mapping factor graph}

\def\nupm{N_{UPM}}
\def\nglobal{N_g}
\def\nevery{N_e}
\def\niter{N_i}

In the absence of trajectory and velocity priors (\emph{no} GPS, \emph{no} odometry, etc.), the factor graph used for state estimation is built iteratively and optimised as new factors are added to the cost function.
Fig.~\ref{algorithm:slam_process} shows the proposed strategy's algorithm.
Intuitively, the first frames need particular attention because the initial state is completely unknown when the system is switched on.
Therefore, during the initialisation step, the integration of any single new frame triggers both the optimisation of the state $\state$, and the feature recomputation and data association for \emph{every} frame already in the graph.
After this initialisation step, motion-distortion is removed from the corresponding lidar scans.
Feature recomputation in these frames is not needed later in the process (the features are reliable as computed on distortion-free scans).
Only the latest frames have their features and data association recomputed as $\state$ changes.
Note that the method still considers motion distortion in \emph{all} the frames at all times because UPMs are used in the lidar residuals, and the full trajectory is part of the state $\state$.
Integrating IN2LAAMA into a more complex system that provides reliable prior information (e.g., robot's odometry, GPS if outdoors, etc.), could significantly reduce the number of iterations needed to build the factor graph (potentially in one go).
The execution time would be greatly reduced.

\begin{figure}
    \hrulefill
    \vspace{-0.1cm}
    \begin{algorithmic}[1]
        \scriptsize
        \renewcommand{\algorithmicrequire}{\textbf{Input:}}
        \renewcommand{\algorithmicensure}{\textbf{Output:}}
        \Require \textbf{ }
        \Statex $\nblidarframe$: Number of frames in the dataset
        \Statex $\nglobal$: Number of frames to initialise initial conditions
        \Statex $\nevery$: Number of frames added between each optimisation
        \Statex $Calib$: Activate calibration parameters estimation
        \Ensure  \textbf{ }
        \Statex $\state$: State estimate
        \Statex
        \State $F \gets$ Create empty factor graph   \hfill \textit{// Start initialisation //}
        \For {$n = 0:\nglobal-1$} \hfill \textit{//}
            \State Add frame $n$ and associated factors to $F$ \hfill \textit{//}
            \Repeat \hfill \textit{//}
                \State $\state \gets Optimise(F)$ \hfill \textit{//}
                \State Check/recompute features in frames 0 to $n$ \hfill \textit{//}
            \Until{ Reach nb. iterations $\parallel$ State converges } \hfill \textit{//}
        \EndFor \hfill \textit{//}
        \If{$Calib$} \hfill \textit{//}
            \State $N \gets 1$ \hfill \textit{//}
        \Else \hfill \textit{//}
            \State $N \gets \nevery$ \hfill \textit{//}
        \EndIf \hfill \textit{// End initialisation //}
        \State
        \For {$n = \nglobal:\nblidarframe-1$}
            \State Add frame $n$ and associated factors to $F$
            \If{$n \mod N = 0$ $\parallel$ $n = \nblidarframe-1$}
                \Repeat
                    \State $\state \gets Optimise(F)$
                    \State Check/recompute features in frames $n-\nevery$ to $n$
                \Until{ Reach nb. iterations $\parallel$ State converges }
            \EndIf
            \State Check loop-closure. If loop detected: $\state \gets Optimise(F)$
        \EndFor
        \If{$Calib$}
            \State Add $\Tc$ to $\state$
            \Repeat
                \State $\state \gets Optimise(F)$
                \State Check/recompute features in frames $0$ to $\nblidarframe-1$
            \Until{ Reach nb. iterations $\parallel$ State converges }
        \EndIf
    \end{algorithmic}
    \vspace{-0.4cm}
    \hrulefill

    \caption{Algorithm of the factor graph construction and optimisation procedure.}
    \label{algorithm:slam_process}
\end{figure}

\subsection{Calibration factor graph}

The localisation and mapping procedure relies on a good knowledge of $\Tc$. Using inaccurate calibration parameters can lead to contradicting information in the factor graph.
As a consequence, using the integration of inertial data from the last estimate of the state provides a prior that drifts rapidly.
For the autocalibration procedure, IN2LAAMA runs the localisation and mapping procedure based on  any prior knowledge of $\Tc$, while performing the optimisation step every time a new frame is added to the factor graph (cf. $N \gets 1$ in the algorithm of Fig.~\ref{algorithm:slam_process}).
Once the full trajectory is estimated based on the inaccurate calibration parameters, the calibration parameters $\Tc$ are included as part of the state $\state$ to be estimated along with the different poses, velocities, and bias and time-shift corrections already in $\state$.
Given this new $\state$, the proposed method iteratively optimises the factor graph and recomputes features until the estimate converges.

\subsection{Robustness of state estimation}

A major challenge for robotics state estimation is the management of outliers.
In a localisation and mapping framework, like the one presented here, outliers can originate through different phenomena.
One is simply wrong data association.
Despite conservative lidar feature-association rules, in cluttered environments, it is still likely that outliers pass the rejection tests mentioned in Section~V-C.
To address this issue, bisquare weights are applied to each individual lidar residual.

Another source of outliers is the ``quality" of the sensor models used (one can interpret it the other way around as ``the quality of the sensor data").
By definition, a mathematical model is an approximation (more or less accurate) used to describe a real-world system.
IN2LAAMA uses common models for the sensor readings (additive zero-mean Gaussian noise) and the IMU biases (Brownian motion).
These considerations might not capture reality accurately.
In a multi-sensor estimation framework, sensor data that do not correspond exactly to the models employed create contradicting information in the estimation process.
We propose to overcome this issue by applying Cauchy loss functions on each of the lidar and IMU factors to attenuate these outliers.
The experiment in VII.B.3 shows the robustness gain in the presence of erroneously modelled IMU measurements.

\subsection{Bias observability}

The inertial navigation literature has previously studied the observability of IMU biases in different estimation frameworks \cite{Tereshkov2013,Yu2016,Du2017}.
It has been proven that in the presence of inertial data, the biases of the accelerometer are observable only if the attitude of the system is perfectly known or if the trajectory contains rotations \cite{Tereshkov2015}.
If none of these conditions is satisfied, the estimated state $\state$ is not unique.
In the case of lidar-inertial fusion, a lidar alone cannot provide an accurate global attitude without relying on strong heuristics that are not desired in generic localisation and mapping frameworks (e.g. Manhattan world with walls aligned with gravity vector).
Therefore, an extra constraint is needed to tackle the problem of translation-only trajectories.
The proposed method overcomes this problem by integrating a simple factor that penalises the distance between the estimated accelerometer biases and the null vector.

This additional factor on $\bacccorrectionz$ is added to the factor graph upon creation.
Once the final frame's factors are added, the observability constraint is released (weight null), and the cost function is minimised.
If the magnitude of the accelerometer biases is far from zero, the constraint is re-established and the optimisation run again.
This strategy covers the scenarios where the estimation ambiguity is present only at the start of the trajectory.
Note that in the non-observable cases, the estimated biases and global orientation are inaccurate, but the trajectory, therefore the map, is still consistent.

\subsection{UPMs and memory}

The main attribute of the UPMs is to make precise inertial data available for each of the points collected by the lidar and therefore allowing for precise motion distortion correction.
The drawbacks of these measurements are the computation time (due to the GP interpolations and the numerical integration) and memory usage.
Because the UPMs are relatively slow to compute, storing them is essential to limit the global execution time.
On the other hand, UPMs need a significant amount of memory as each UPM is stored on at least 150 floating-point numbers (preintegrated measurements, covariance matrix, Jacobians for bias and time-shift corrections).
Based on the Velodyne VLP-16 and double-precision floating-point numbers, it represents a memory consumption of more than 340MB per second of data solely to store the UPMs.
To reduce the memory footprint and make IN2LAAMA executable on standard computers, only the UPMs associated to the last $\nupm$ frames are stored.
As per the localisation and mapping procedure shown in the algorithm of Fig.~\ref{algorithm:slam_process}, choosing $\nupm$ equal to $\max(\nglobal, \nevery)$ does not impact the estimation time.
However, this strategy requires the recomputation of all the UPMs to export the dense map when the estimation process terminates.

\section{Experiments and results}

The proposed framework has been evaluated in simulation and on real data from our platform and a public dataset.
Our real-world platform is a self-contained lidar-inertial system:
\begin{itemize}
    \item Velodyne VLP-16, 16-channel ($\pm 15\,{}^\circ$) lidar rotating at $10\,\mathrm{Hz}$ providing 300k points per second (noise of $\pm3\,\mathrm{cm}$).
    \item Xsens MTi-3, 3-axis accelerometer and 3-axis gyroscope sampling at 100Hz (noise of $0.02\,\mathrm{m/s^2}$ and $0.097\,\mathrm{{}^\circ/s}$).
\end{itemize}
The simulated datasets have been generated to match the characteristics of the above-mentioned system moving in a virtual room constituted of 7 planes.
Both the back-end and front-end of the proposed method are tested and evaluated in our simulated experiments.
In the rest of this section, the A-LOAM implementation\footnote{https://github.com/HKUST-Aerial-Robotics/A-LOAM} of \cite{Zhang2014b} is used to benchmark the proposed method.
This last technique has been chosen for its top performance with lidar systems in the KITTI odometry benchmark \cite{Geiger2012}.
Our implementation of IN2LAAMA is built upon the non-linear least-square solver Ceres\footnote{http://ceres-solver.org} with analytical Jacobians of the cost function.

\subsection{Simulation - UPM}
This section discusses the accuracy of GP-based UPMs compared to other preintegration methods.
The original preintegration method presented in \cite{Lupton2012} and \cite{Forster2015a} relies on the hypothesis that acceleration and angular velocity measurements are constant between two consecutive IMU sampling time.
This creates integration noise.
The UPMs, introduced in \cite{LeGentil2018}, first upsample the inertial measurements before performing numerical integration, thus improving the accuracy of the preintegrated measurements.

\begin{figure}
    \centering
	\begin{tikzpicture}
        \def\plotwidth{3.55cm}
        \def\plotheight{2.9cm}
		\begin{axis}[
                xlabel={\tiny IMU frequency (Hz)},
                ylabel={\tiny RMSE pos. error (m)},
            xmin=25, xmax=100,
            width=\plotwidth,
            height=\plotheight,
            ymin=9e-6, ymax=0.02,
            ymode=log,
            xtick={25,50,75,100},
			every tick label/.append style={font=\fontsize{4}{4}},
            y label style={at={(-0.30,0.4)}},
            x label style={at={(0.5,-0.12)}}
			]
			\addplot[
			color=red
			]
			coordinates {
                (25,7.1279e-05)
                (33.3,5.0953e-05)
                (50, 5.1029e-05)
                (100, 3.4545e-05)
			};
			\addplot[
			color=orange
			]
			coordinates {
                (25, 8.897e-05)
                (33.3, 6.3352e-05)
                (50, 6.7101e-05)
                (100, 4.7083e-05)
			};
			\addplot[
			color=greentam
			]
			coordinates {
                (25, 0.0007248)
                (33.3, 0.00048326)
                (50, 0.00037077)
                (100, 0.00044018)
			};
			\addplot[
			color=bluetam
			]
			coordinates {
                (25, 0.015637)
                (33.3, 0.012529)
                (50, 0.0081658)
                (100, 0.0053458)
			};
        \end{axis}
	\end{tikzpicture}
	\begin{tikzpicture}
        \def\plotwidth{3.55cm}
        \def\plotheight{2.9cm}
		\begin{axis}[
                xlabel={\tiny IMU frequency (Hz)},
                ylabel={\tiny RMSE rot. error (${}^\circ$)},
            xmin=25, xmax=100,
            width=\plotwidth,
            height=\plotheight,
            ymin=9e-4, ymax=2,
            ymode=log,
            xtick={25,50,75,100},
			every tick label/.append style={font=\fontsize{4}{4}},
            y label style={at={(-0.30,0.4)}},
            x label style={at={(0.5,-0.12)}}
			]
			\addplot[
			color=red
			]
			coordinates {
                (25, 0.005735)
                (33.3, 0.00500572)
                (50, 0.003870287)
                (100, 0.00318)
			};
			\addplot[
			color=orange
			]
			coordinates {
                (25, 0.007086)
                (33.3, 0.00666)
                (50, 0.005235248)
                (100, 0.00412)
			};
			\addplot[
			color=greentam
			]
			coordinates {
                (25, 0.01177)
                (33.3, 0.008869299)
                (50, 0.006929427)
                (100, 0.00466)
			};
			\addplot[
			color=bluetam
			]
			coordinates {
                (25, 0.4525)
                (33.3, 0.3496)
                (50, 0.228731847)
                (100, 0.121)
			};
        \end{axis}
	\end{tikzpicture}
    \begin{tikzpicture}[baseline=0pt]
        \def\lengthline{0.2}
        \coordinate(upmgp)   at (0  , 0.28);
        \coordinate(upmlin)  at (0  , 0.81);
        \coordinate(upmhold) at (0  , 1.34);
        \coordinate(pm)      at (0  , 1.87);
        \coordinate[right = \lengthline of upmgp]  (bisupmgp)   ;
        \coordinate[right = \lengthline of upmlin] (bisupmlin)  ;
        \coordinate[right = \lengthline of upmhold](bisupmhold) ;
        \coordinate[right = \lengthline of pm]     (bispm)      ;
        \draw[red] (upmgp) -- (bisupmgp);
        \draw[orange] (upmlin) -- (bisupmlin);
        \draw[greentam] (upmhold) -- (bisupmhold);
        \draw[bluetam] (pm) -- (bispm);
        \node[execute at begin node=\setlength{\baselineskip}{6pt}, text width = 1.6cm, right = \lengthline of upmgp]  {\scriptsize UPM (GP regression)};
        \node[execute at begin node=\setlength{\baselineskip}{6pt}, text width = 1.6cm, right = \lengthline of upmlin] {\scriptsize UPM (linear interpolation)};
        \node[execute at begin node=\setlength{\baselineskip}{6pt}, text width = 1.6cm, right = \lengthline of upmhold]{\scriptsize UPM (zero-order hold)};
        \node[execute at begin node=\setlength{\baselineskip}{6pt}, text width = 1.6cm, right = \lengthline of pm]     {\scriptsize \cite{Zhang2014b}};

	\end{tikzpicture}
    \caption{Accuracy of different preintegration methods as a function of the IMU frequency. Pose RMSE at the lidar points' timestamps (averaged over a 50-run Monte Carlo simulation).}
	\label{figure:upm}
\end{figure}
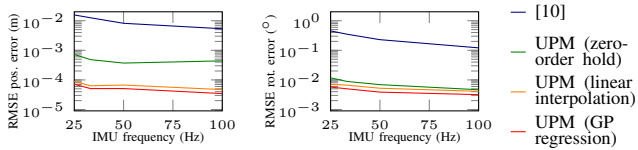

This set-up compares the Root Mean Square Error (RMSE) of the preintegrated measurements computed for each of the 3D-points of a lidar scan using the classic preintegration from \cite{Forster2015a} at IMU frequency, and the UPMs based on different interpolation methods: zero-order hold, linear, and GP regression.
As shown in Fig.~\ref{figure:upm}, all versions of UPMs outperform the standard preintegrated measurements by at least an order of magnitude.
Among UPM methods, the accuracy varies slightly depending on the chosen interpolation method.
Additionally, a 3D animation demonstrating the different preintegration methods over a long integration window is depicted in the supplementary video.
As per its non-parametric and probabilistic characteristics, GPs offer a better description of the underlying signal compared to simple linear interpolations.
Consequently, UPMs based on GP regression performs best among the benchmarked methods.
In the rest of the paper, we employ the more principled GP interpolation, but one can use linear interpolation to trade a slight loss of accuracy for lower computation time (in our implementation, it represents a gain of approximatively $7\,\mathrm{s}$ of computation per second of data).  

\begin{table*}
    \caption{Quantitative results of the odometry set-up in simulated environment (50-run Monte Carlo simulation).}
    \begin{center}
    \vspace{-0.2cm}
    \newcolumntype{C}[1]{>{\centering\let\newline\\\arraybackslash\hspace{0pt}}m{#1}}
    \def\colsize{1.60cm}
    \def\largecolsize{2.2cm}
    \def\textsize{\scriptsize}
    \def\smallcolsize{0.65cm}
    \begin{tabular}{C{1.80cm}  C{\largecolsize}  C{\smallcolsize}  C{\colsize}  C{\colsize}  C{\colsize}  C{\colsize}  C{\colsize}  C{\colsize} }
        \textsize Motion type (ang. vel. in $\mathrm{\circ/s}$)
        &\textsize Framework
        &\textsize Num. fails
        &\textsize Final pos. error (m)
        &\textsize Final rot. error (${}^\circ$)
        &\textsize Relative pos. error (m)
        &\textsize Relative rot. error (${}^\circ$)
        &\textsize RMSE pos. error (m) & RMSE rot. error (${}^\circ$)
        \\
        \hline
        \footnotesize
        \cellcolor[gray]{.95}
        &\cellcolor[gray]{.95}\textsize \cite{Zhang2014b}
        &\cellcolor[gray]{.95}\textsize \textbf{0}
        &\cellcolor[gray]{.95}\textsize 5.67 $\pm$ 2.63
        &\cellcolor[gray]{.95}\textsize 27.9 $\pm$ 13.6
        &\cellcolor[gray]{.95}\textsize 0.47 $\pm$ 0.14
        &\cellcolor[gray]{.95}\textsize 1.46 $\pm$ 0.51
        &\cellcolor[gray]{.95}\textsize 5.62 $\pm$ 1.72
        &\cellcolor[gray]{.95}\textsize 29.2 $\pm$ 8.98
        \\
        \cellcolor[gray]{.95}
        &\cellcolor[gray]{.95}\textsize \cite{LeGentil2019} (No IMU factors)
        &\cellcolor[gray]{.95}\textsize \textbf{0}
        &\cellcolor[gray]{.95}\textsize 0.11 $\pm$ 0.06
        &\cellcolor[gray]{.95}\textsize 0.80 $\pm$ 0.42
        &\cellcolor[gray]{.95}\textsize 0.01 $\pm$ 0.002
        &\cellcolor[gray]{.95}\textsize 0.03 $\pm$ 0.002
        &\cellcolor[gray]{.95}\textsize 0.08 $\pm$ 0.03
        &\cellcolor[gray]{.95}\textsize 0.56 $\pm$ 0.24
        \\
        \multirow{-3}{\colsize}{\cellcolor[gray]{.95}\centering\textsize Slow (avg~14.7, max~22.1)}
        &\cellcolor[gray]{.95}\textsize IN2LAAMA
        &\cellcolor[gray]{.95}\textsize \textbf{0}
        &\cellcolor[gray]{.95}\textsize \textbf{0.06} $\pm$ 0.03
        &\cellcolor[gray]{.95}\textsize \textbf{0.12} $\pm$ 0.07
        &\cellcolor[gray]{.95}\textsize \textbf{0.003} $\pm$ 3e-4
        &\cellcolor[gray]{.95}\textsize \textbf{0.005} $\pm$ 1e-4
        &\cellcolor[gray]{.95}\textsize \textbf{0.04} $\pm$ 0.02
        &\cellcolor[gray]{.95}\textsize \textbf{0.09} $\pm$ 0.04
        \\
        \hline
        & \textsize \cite{Zhang2014b}
        & \textsize \textbf{0}
        & \textsize 7.03 $\pm$ 3.41
        & \textsize 57.2 $\pm$ 26.2
        & \textsize 0.48 $\pm$ 0.17
        & \textsize 4.85 $\pm$ 1.84
        & \textsize 6.29 $\pm$ 2.22
        & \textsize 56.1 $\pm$ 17.9
        \\
        & \textsize\cite{LeGentil2019} (No IMU factors)
        & \textsize 1
        & \textsize 0.34 $\pm$ 0.26
        & \textsize 2.17 $\pm$ 1.51
        & \textsize 0.02 $\pm$ 0.004
        & \textsize 0.06 $\pm$ 0.009
        & \textsize 0.24 $\pm$ 0.12
        & \textsize 1.70 $\pm$ 0.86
        \\
        \multirow{-3}{\colsize}{\centering\textsize Moderate (avg~49.0, max~78.2)}
        & \textsize IN2LAAMA
        & \textsize \textbf{0}
        & \textsize \textbf{0.30} $\pm$ 0.57
        & \textsize \textbf{1.37} $\pm$ 6.31
        & \textsize \textbf{0.004} $\pm$ 0.002
        & \textsize \textbf{0.009} $\pm$ 0.018
        & \textsize \textbf{0.19} $\pm$ 0.38
        & \textsize \textbf{0.81} $\pm$ 3.63
        \\
        \hline
        \cellcolor[gray]{.95}
        &\cellcolor[gray]{.95}\textsize \cite{Zhang2014b}
        &\cellcolor[gray]{.95}\textsize \textbf{0}
        &\cellcolor[gray]{.95}\textsize 16.2 $\pm$ 5.60
        &\cellcolor[gray]{.95}\textsize 119 $\pm$ 38.2
        &\cellcolor[gray]{.95}\textsize 0.53 $\pm$ 0.12
        &\cellcolor[gray]{.95}\textsize 13.0 $\pm$ 3.45
        &\cellcolor[gray]{.95}\textsize 11.1 $\pm$ 2.71
        &\cellcolor[gray]{.95}\textsize 90.7 $\pm$ 20.6
        \\
        \cellcolor[gray]{.95}
        &\cellcolor[gray]{.95}\textsize \cite{LeGentil2019} (No IMU factors)
        &\cellcolor[gray]{.95}\textsize 37
        &\cellcolor[gray]{.95}\textsize \textbf{0.39} $\pm$ 0.14
        &\cellcolor[gray]{.95}\textsize 3.39 $\pm$ 1.65
        &\cellcolor[gray]{.95}\textsize 0.04 $\pm$ 0.01
        &\cellcolor[gray]{.95}\textsize 0.10 $\pm$ 0.02
        &\cellcolor[gray]{.95}\textsize \textbf{0.28} $\pm$ 0.09
        &\cellcolor[gray]{.95}\textsize 2.38 $\pm$ 0.92
        \\
        \multirow{-3}{\colsize}{\cellcolor[gray]{.95}\centering\textsize Fast (avg~125, max~198)}
        &\cellcolor[gray]{.95}\textsize IN2LAAMA
        &\cellcolor[gray]{.95}\textsize \textbf{0}
        &\cellcolor[gray]{.95}\textsize 0.96 $\pm$ 0.61
        &\cellcolor[gray]{.95}\textsize \textbf{0.59} $\pm$ 3.00
        &\cellcolor[gray]{.95}\textsize \textbf{0.007} $\pm$ 0.003
        &\cellcolor[gray]{.95}\textsize \textbf{0.007} $\pm$ 0.009
        &\cellcolor[gray]{.95}\textsize 0.57 $\pm$ 0.36
        &\cellcolor[gray]{.95}\textsize \textbf{0.36} $\pm$ 1.71
    \end{tabular}
    \label{table:imu_factor}
    \end{center}
    \vspace{-0.1cm}
    {\footnotesize  The trajectories have an average length of $288.7\,\mathrm{m}$, an average velocity of $4.85\,\mathrm{m/s}$, and a maximum velocity of $7.35\,\mathrm{m/s}$.  The different rows correspond to different levels of angular velocity during the trajectory. The errors displayed are computed only on the successful runs against the ground truth. The RMSE errors are computed all along each trajectory estimates. The relative errors correspond to frame-to-frame registration errors.}
\end{table*}
\subsection{Simulation - Localisation and mapping}

This set-up aims at evaluating different configurations of the proposed framework for localisation and mapping.
The simulated trajectories have been generated from sine functions with random frequencies and amplitudes.
The extrinsic calibration between sensors is randomly generated for each trajectory.
The results are evaluated over 50-run Monte Carlo simulations.

\subsubsection{Odometry}

First, we want to evaluate the advantages of using IMU factors for odometry-like localisation and mapping by comparing the accuracy of IN2LAAMA against \cite{Zhang2014b}, and our previous work \cite{LeGentil2019}.
We evaluate the localisation accuracy of the three frameworks on three sets of trajectories that have different levels of angular velocities.
The loop closure detection is deactivated for these experiments.

Table~\ref{table:imu_factor} reports the trajectories' parameters as well as the localisation errors against ground truth for \cite{Zhang2014b}, \cite{LeGentil2019}, and IN2LAAMA.
The poor performance of \cite{Zhang2014b} is easily explained by the fact that the motion starts at the very beginning of the simulation.
Consequently, even the first lidar frame contains motion distortion.
This cannot be properly corrected by \cite{Zhang2014b} and leads to unrecoverably large localisation error.
In the case of low angular velocities (row ``Slow"), both \cite{LeGentil2019} and IN2LAAMA succeed in estimating the system pose.
The integration of IMU factors improves the trajectory estimate slightly.
Note that the presence of higher angular velocities leads to smaller overlap between lidar scans.
Thus, the ``Moderate" and ``Fast" trajectories contain cases where the overlap between consecutive scans is not enough for the lidar factors to fully constrain the frame-to-frame motion.
These ``degenerated" trajectories trigger estimation failure of \cite{LeGentil2019} but are correctly handled by the integration of IMU factors.
Note that the accuracy metrics of Table~\ref{table:imu_factor} are computed on successful runs only, therefore advantaging \cite{LeGentil2019} in the ``Fast" scenario.

\subsubsection{Loop-closure}

This set-up aims at demonstrating the ability of the proposed method to perform simultaneous localisation and mapping (SLAM) by integrating loop closures in the batch optimisation.
A set of simulated trajectories is generated so that the first and last poses coincide.
Table~\ref{table_loop_closure} shows the localisation results with and without the proposed loop closure detection method.
The numbers show that loop closures help to reduce both the final pose error as well as the overall localisation error all along the trajectory. 

\begin{table}
    \caption{Localisation accuracy with/without loop closures.}
    \begin{center}
    \vspace{-0.2cm}
    \newcolumntype{C}[1]{>{\centering\let\newline\\\arraybackslash\hspace{0pt}}m{#1}}
    \def\colsize{2.3cm}
    \def\textsize{\scriptsize}
        \begin{tabular}{C{\colsize}| C{\colsize} C{\colsize}}
            \textsize Loop-closure
            & \textsize Final pose error
            & \textsize RMSE pose error 
            \\
            \hline
            \cellcolor[gray]{.95}&
            \cellcolor[gray]{.95}\textsize 0.110$\,\mathrm{m}\,\pm$ 0.043&
            \cellcolor[gray]{.95}\textsize 0.051$\,\mathrm{m}\,\pm$ 0.021
            \\
            \cellcolor[gray]{.95}\multirow{-2}{\colsize}{\centering\textsize Without}&
            \cellcolor[gray]{.95}\textsize 0.30$\,\mathrm{{}^\circ}\,\pm$ 0.19&
            \cellcolor[gray]{.95}\textsize 0.17$\,\mathrm{{}^\circ}\,\pm$ 0.11
            \\
            \hline
            &
            \textsize \textbf{0.011}$\,\mathrm{m}\,\pm$ 0.011&
            \textsize \textbf{0.019}$\,\mathrm{m}\,\pm$ 0.007 
            \\
            \multirow{-2}{\colsize}{\centering\textsize With}&
            \textsize \textbf{0.15}$\,\mathrm{{}^\circ}\,\pm$ 0.12&
            \textsize \textbf{0.10}$\,\mathrm{{}^\circ}\,\pm$ 0.07
        \end{tabular}
    \end{center}
    \label{table_loop_closure}
    \vspace{-0.1cm}
    {\footnotesize The set of 50 simulated trajectories have a mean distance of $210\,\mathrm{m}$, mean velocity of $3.53\,\mathrm{m/s}$, and mean angular velocity of $8.16\,\mathrm{{}^\circ/s}$. The relative errors correspond to frame-to-frame registration errors.}
\end{table}

\subsubsection{Robustness to inaccurate sensor model}

This set-up has been designed to demonstrate the gain of robustness brought by the use of Cauchy loss functions on lidar and IMU factors.
As mentioned in Section~VI-C, a sensor model is only an approximation of reality.
In real data, the sensors do not always behave as per the manufacturer specifications or noise models.
This is the scenario that we are trying to emulate in this set-up.

To simulate a ``mismatch" between model and reality, we perturb the IMU sensitivity by simply multiplying the inertial data by a constant.
Looking at the localisation accuracy shown in Table~\ref{table:bad_imu}, both in terms of error and number of failures, it is clear that the loss functions make IN2LAAMA more robust to discrepancies between the IMU model and the actual readings.

\begin{table}
    \caption{Localisation accuracy with/without Cauchy loss functions.}
    \begin{center}
    \vspace{-0.2cm}
    \newcolumntype{C}[1]{>{\centering\let\newline\\\arraybackslash\hspace{0pt}}m{#1}}
    \def\colsize{1.5cm}
    \def\colsizebis{1.8cm}
    \def\textsize{\scriptsize}
        \begin{tabular}{C{\colsize} | C{\colsizebis} C{\colsizebis} C{\colsizebis}}
            \multirow{2}{\colsize}{\centering\textsize Cauchy loss} & \multicolumn{3}{c}{\tiny{IMU data multipliers}}
            \\
            & \textsize 1.01 & \textsize 1.03 & \textsize 1.05
            \\
            \hline
            \rowcolor[gray]{0.95}
            \textsize Without &
            \textsize 0.23 $\pm$ 0.09 (0) &
            \textsize 0.70 $\pm$ 0.27 (12)&
            \textsize 1.03 $\pm$ 0.31 (34)
            \\
            \hline
            \textsize With &
            \textsize \textbf{0.21} $\pm$ 0.14 (0) &
            \textsize \textbf{0.51} $\pm$ 0.19 (0) &
            \textsize \textbf{0.80} $\pm$ 0.29 (0)
        \end{tabular}
    \end{center}
    \vspace{-0.1cm}
    {\footnotesize The results represent the RMSE position error, in meters, computed upon a 50-run Monte Carlo simulation. The digits in parenthesis are the numbers of failure cases. The trajectories are $47.2\,\mathrm{m}$-long on average and have the ``Fast" velocity profile (linear and angular) from Table~\ref{table:imu_factor}.}
    \label{table:bad_imu}
\end{table}

\subsubsection{No motion model}
Using individual preintegrated measurements for each lidar point, while leveraging non-parametric interpolation, allows the proposed framework to alleviate the constraints of an explicit motion model.
This set-up aims at demonstrating the importance of not imposing a motion model to the state estimation.
We compare IN2LAAMA with a modified version of it built on the assumption of constant angular and linear velocities during the lidar frames.
Table~\ref{table:constant_vel} displays both frameworks' pose errors over a 50-run simulation.
The results clearly demonstrate the accuracy gain of alleviating the use of such a restrictive motion model.

\begin{table}
    \caption{Localisation accuracy with/without motion model.}
    \begin{center}
    \vspace{-0.2cm}
    \newcolumntype{C}[1]{>{\centering\let\newline\\\arraybackslash\hspace{0pt}}m{#1}}
    \def\colsize{2.2cm}
    \def\colsizebis{1.06cm}
    \def\colsizeter{1.5cm}
    \def\textsize{\scriptsize}
        \begin{tabular}{C{\colsizeter}| C{\colsizebis} C{\colsize} C{\colsize}}
            &\textsize Num. fails
            &\textsize RMSE pos. error (m)
            &\textsize RMSE rot. error (${}^\circ$)
            \\
            \hline
            \rowcolor[gray]{0.95}
            \textsize Constant vel. &
            \textsize 18 &
            \textsize 2.67 $\pm$ 4.84 &
            \textsize 19.9 $\pm$ 24.7
            \\
            \hline
            \textsize IN2LAAMA{ \tiny(no motion model)} &
            \textsize \textbf{0} &
            \textsize \textbf{0.087} $\pm$ 0.041 &
            \textsize \textbf{0.088} $\pm$ 0.267
        \end{tabular}
    \end{center}
    \vspace{-0.1cm}
    {\footnotesize IN2LAAMA is compared with a modified version that assumes constant angular and linear velocities. Results are obtained over a 50-run Monte Carlo simulation. The trajectories are $95.2\,\mathrm{m}$-long on average and have the ``Fast" velocity profile (linear and angular) from Table~\ref{table:imu_factor}.}
    \label{table:constant_vel}
\end{table}

\subsection{Simulation - Front-end}

This subsection discusses the proposed feature extraction and compares it with the front-end of \cite{Zhang2014b}. 
We show that our method has a consistent behaviour with respect to the observed surface by using a scoring system robust to the lidar viewpoint.
In this regard, we have computed the feature score for each of the points of simulated scans.
Our method computes a score that represents the cosine of the angle present in a patch.
The higher the score (with 1 being the maximum value), the more planar the patch.
In \cite{Zhang2014b}, the point score gives an evaluation of smoothness in a patch but does not correspond to any particular physical measurement of the actual geometry.
Nonetheless, this score tends to be low in planar patches and high in edge-like patches.

\begin{figure}
	\centering
    \def\imageheight{1.4}
    \def\zoomheight{0.9}
    \newcommand\piece[7]{
        \node[anchor=south,inner sep=0,black]    at (#1-0.7,#2+0.90) {\scriptsize #7};
        \node[anchor=south,inner sep=0,black!70] at (#1-0.7,#2+0.60) {\scriptsize Pose #6};
        \node[anchor=south west,inner sep=0] at (#1+0,#2+0) {\includegraphics[clip, trim=5.5cm 12cm 6cm 12.3cm, height=\imageheight cm]{images/#3_hist}};
        \node[anchor=south west,inner sep=0] at (#1+0.20+#4,#2+0.35) {\includegraphics[clip, trim=6.5cm 13cm 7.5cm 13cm, height=\zoomheight cm]{images/#3_hist_zoom}};
        \node at (#1+0.53+#4+#5,#2+0.81) [anchor=south west,draw=black, fill=white]{\tiny Zoom};
        \node[anchor=south west,inner sep=0] at (#1+3,#2+0) {\includegraphics[clip, trim=3.5cm 10cm 5cm 10cm, height=\imageheight cm]{images/#3_pc_score}};
        \node[anchor=south west,inner sep=0] at (#1+5.2,#2+0) {\includegraphics[clip, trim=3.5cm 10cm 5cm 10cm, height=\imageheight cm]{images/#3_pc_features}};
    }
    \begin{tikzpicture}
        \piece{0}{2*\imageheight+0.2}{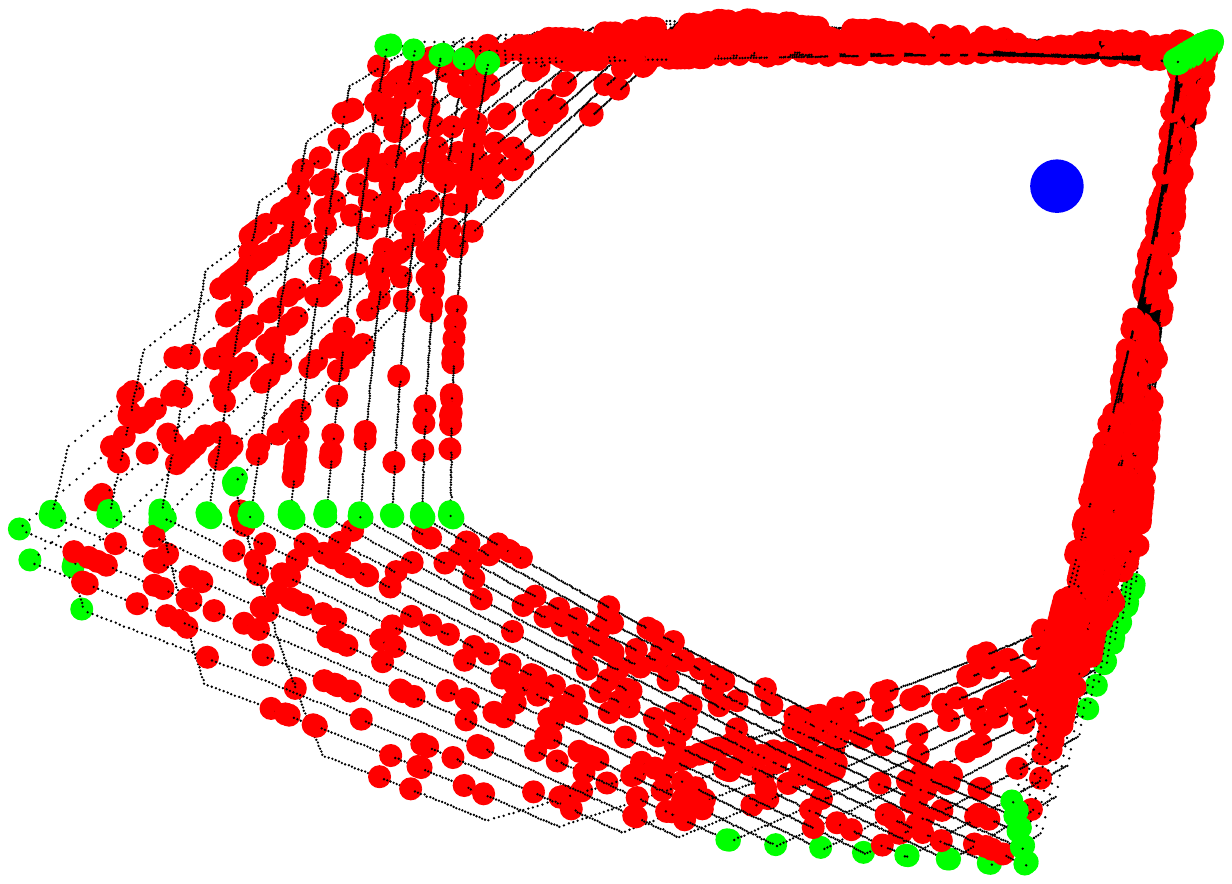}{0}{0}{B}{IN2LAAMA}
        \piece{0}{3*\imageheight+0.3}{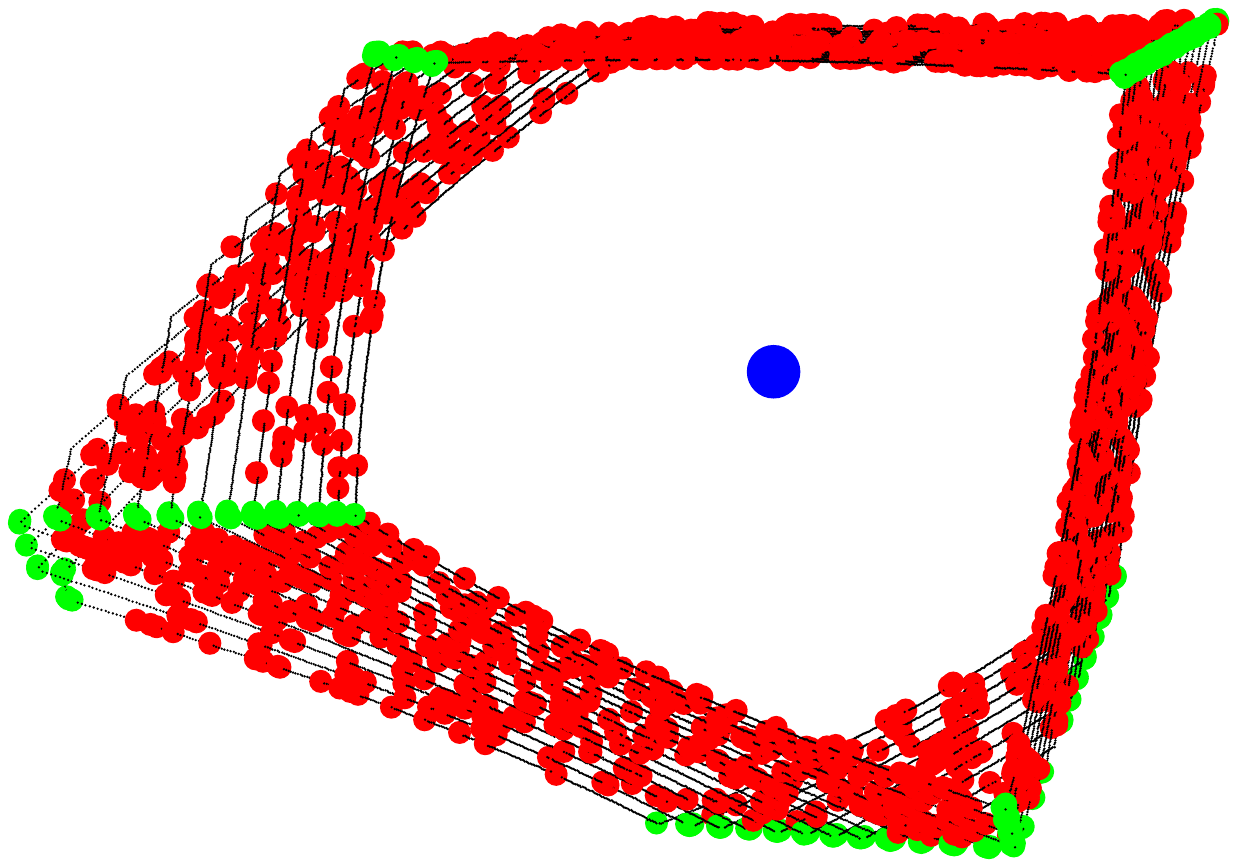}{0}{0}{A}{IN2LAAMA}
        \piece{0}{0}{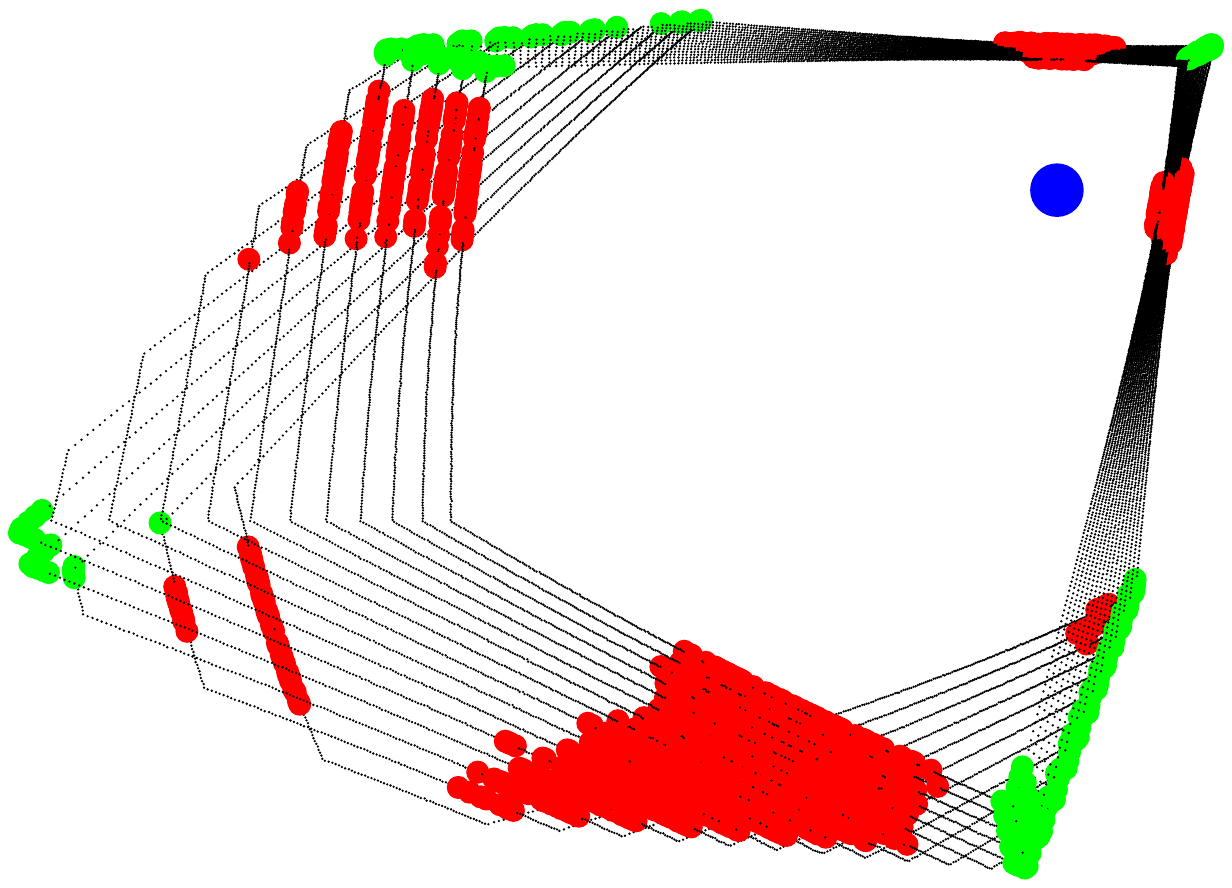}{0.45}{0.69}{B}{\cite{Zhang2014b}}
        \piece{0}{\imageheight+0.1}{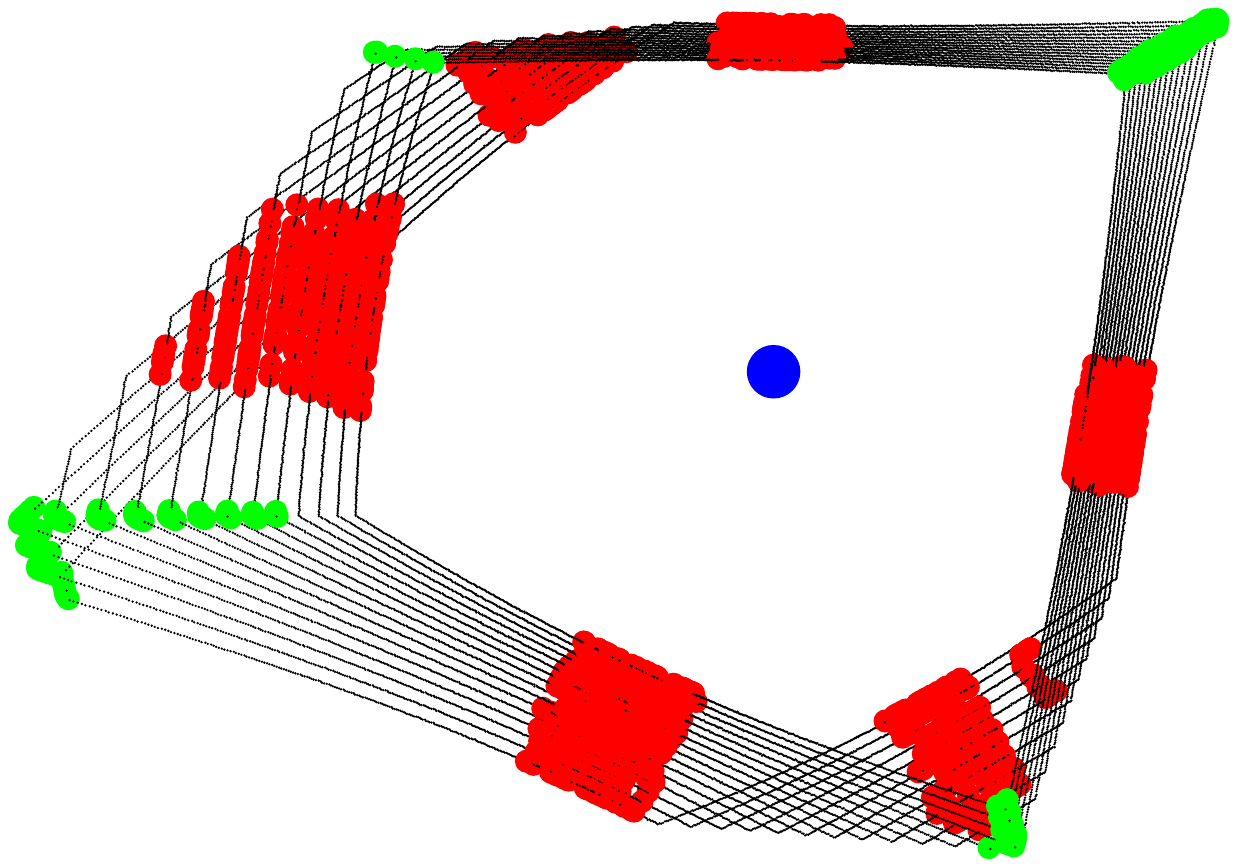}{0.45}{0.69}{A}{\cite{Zhang2014b}}
        \draw[black] (-1.4,3*\imageheight+0.3) -- (7.2,3*\imageheight+0.3);
        \draw[black] (-1.4,\imageheight+0.1) -- (7.2,\imageheight+0.1);
        \draw (0,0) -- (0,4*\imageheight+0.3);
        \draw (2.9,0.0) -- (2.9,4*\imageheight+0.3);
        \draw (5.1,0.0) -- (5.1,4*\imageheight+0.3);
        \draw (-1.4,2*\imageheight+0.2) -- (7.2,2*\imageheight+0.2);
        \node[anchor=south,inner sep=0,black] at (1.4,-0.25) {\footnotesize (a) Histogram};
        \node[anchor=south,inner sep=0,black] at (3.95,-0.25) {\footnotesize (b) Score};
        \node[anchor=south,inner sep=0,black] at (6.15,-0.25) {\footnotesize (c) Features};
    \end{tikzpicture}
    \caption{Feature score comparison between IN2LAAMA and \cite{Zhang2014b}. (a) Histograms of the points' score. (b): Spatial visualisation of the score (same colours as histograms). (c): Features selected with the 100 ``most planar" points of each lidar channel in red and the 15 sharpest edge points in green. IN2LAAMA's score exposes a physical value (here, edges have a score $\cos(\beta) < \cos(45^\circ)$, cf. Fig.~\ref{figure:feature_extraction}\,(b)). The blue dots are the lidar position.}
	\label{figure:feature_score}
\end{figure}

Fig.~\ref{figure:feature_score}\,(a) shows the histograms of the scores in our simulated environment.
It illustrates that the scores computed by IN2LAAMA are consistent across the scan (the bin around 1 dominates the histograms largely).
The zoomed-in plots show the modes associated with the different edges of the environment.
However, with \cite{Zhang2014b}'s scoring system (inverse to ours), the distinction of the different surface types is ambiguous.
Fig.~\ref{figure:feature_score}\,(b) spatially represents the points' score with colours and Fig.~\ref{figure:feature_score}\,(c) shows the subsequent feature selection.
One can see that IN2LAAMA's score is consistent with the observed surface despite changes in the lidar's pose.
Having features spread all across the scene leads to better estimation stability.

To evaluate this last point, we compare the proposed method (front-end and back-end) against a hybrid method that uses our back-end in association with the front-end of \cite{Zhang2014b} in the simulated environment.
The proposed method leads to an RMSE pose error of $0.087\,\mathrm{m}$ and $0.088\,\mathrm{{}^\circ}$, while the hybrid method resulted in an error of $0.431\,\mathrm{m}$ and $0.220\,\mathrm{{}^\circ}$.
These numbers have been computed over a 50-run Monte Carlo simulation.
The average trajectory length is $95.2\,\mathrm{m}$, the average velocity $4.86\,\mathrm{m/s}$, and the average angular velocity $125\,\mathrm{{}^\circ/s}$.

\subsection{Simulation - Calibration}

This set-up aims to evaluate the accuracy of the proposed method when used for extrinsic calibration between a lidar and an IMU.
These series of experiments were run with different error levels of initial guess.
Table~\ref{table_simulated_calibration} shows the error of the calibration estimates.
We can see that the estimates' errors stay small despite an increasingly bad initial guess.
The scenario with the worst initial guess (fourth row of Table~\ref{table_simulated_calibration}) leads to an error slightly bigger than in the three other scenarios.
Nonetheless, in such a case, it is possible to run a second iteration using the first iteration's calibration estimate as the initial guess.
Doing so, the worst initial guess scenario leads to final calibration errors of $10.8\text{e-}3\,\mathrm{m}$, and $0.031\,\mathrm{{}^\circ}$ after the second iteration of IN2LAAMA.
Experiments have also been conducted using longer data recordings, $39.2\,\mathrm{s}$, with the same initial guess as in the first row of Table~\ref{table_simulated_calibration}.
The average errors over 50 runs are reduced to $7.6\text{e-}3\,\mathrm{m}$, and $0.024\,\mathrm{{}^\circ}$.
These results demonstrate similar accuracy as our previous work \cite{LeGentil2018} that relies on observing a calibration target made of at least three non-coplanar planes.

\begin{table}
    \caption{Lidar-IMU extrinsic calibration accuracy.}
    \begin{center}
    \vspace{-0.2cm}
    \newcolumntype{C}[1]{>{\centering\let\newline\\\arraybackslash\hspace{0pt}}m{#1}}
    \def\colsize{2.1cm}
    \def\smallcolsize{2.5cm}
    \def\textsize{\scriptsize}
        \begin{tabular}{C{\smallcolsize}| C{\colsize} C{\colsize}}
            \textsize Avg. error initial guess &
            \textsize Translation error (m) &
            \textsize Rotation error (${}^\circ$)
            \\
            \hline
            \rowcolor[gray]{0.95}
            \textsize $0.17\,\mathrm{m}$, $1.74\,\mathrm{{}^\circ}$ &
            \textsize 10.5e-3 $\pm$ 6.34e-3 &
            \textsize 0.035 $\pm$ 0.037 
            \\
            \hline
            \textsize $0.52\,\mathrm{m}$, $3.47\,\mathrm{{}^\circ}$ &
            \textsize 11.2e-3 $\pm$ 6.94e-3 &
            \textsize 0.047$\pm$ 0.053
            \\
            \hline
            \rowcolor[gray]{0.95}
            \textsize $0.86\,\mathrm{m}$, $8.55\,\mathrm{{}^\circ}$ &
            \textsize 10.8e-3 $\pm$ 6.92e-3 &
            \textsize 0.062 $\pm$ 0.142
            \\
            \hline
            \textsize $1.22\,\mathrm{m}$, $25.1\,\mathrm{{}^\circ}$ &
            \textsize 16.3e-3 $\pm$ 18.7e-3 &
            \textsize 0.125 $\pm$ 0.326
        \end{tabular}
    \end{center}
    \vspace{-0.1cm}
    {\footnotesize Analysis performed over a 50-run Monte Carlo simulation. The different rows correspond to different levels of error on the initial guess. The trajectories used last $19.6\,\mathrm{s}$ ($95.2\,\mathrm{m}$-long on average), and have the same velocity characteristics as the ``Fast" trajectories of our odometry set-up (Table~\ref{table:imu_factor}).}
    \label{table_simulated_calibration}
\end{table}

\subsection{Real-data - Localisation and mapping}

Multiple datasets have been collected with our sensor suite inside the facilities of the University of Technology Sydney.
These datasets contain per-point timestamps and have been made publicly available\footnote{https://github.com/UTS-CAS/in2laama\_datasets}.
Moreover, to demonstrate the versatility of the proposed method, we have applied IN2LAAMA to the MC2SLAM dataset \cite{Neuhaus2018}.
It has been selected because it contains per-lidar-point timestamps.

As mentioned above, our sensor suite comprises a \textit{Velodyne VLP-16} and a low-cost \textit{Xsens MTi-3} IMU.
The \textit{snark} and ROS Xsens drivers\footnote{https://github.com/acfr/snark (lidar) http://wiki.ros.org/xsens\_driver (IMU)} were used to collect the lidar and IMU data, respectively.
Lidar points and IMU measurements were logged with their associated timestamps.
There is no explicit hardware or software mechanism for inter-sensor time synchronisation.

For quantitative comparisons, we chose to use metrics related to the planarity of planes in the environment as per none of the datasets used contains ground truth.
When available, we overlay the map generated with the building's blueprints or area's satellite images.
In the presence of loop closure, we compute the amount of drift without the loop-closure.
A video that shows 3D animations of the maps generated in this subsection is attached to this manuscript.

\subsubsection{Indoors}

This set-up aims to benchmark IN2LAAMA against our previous work \cite{LeGentil2019} (no IMU factors) and the method in \cite{Zhang2014b}.
Real indoor datasets from two different locations have been used for this evaluation: a lab environment and a staircase between floors.
In both cases, we show that the proposed method outperforms both \cite{LeGentil2019} and \cite{Zhang2014b}.

\begin{figure*}
    \def\heightdigit{3.7}
    \def\heightsize{\heightdigit cm}
    \centering
    \newcommand\loam[2]{
        \node[anchor=south west,inner sep=0] at (#1+0,#2+0) {\includegraphics[clip, height=\heightsize]{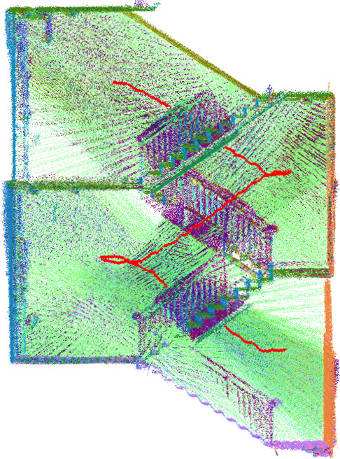}};
        \node[anchor=south,inner sep=0] at (#1+4,#2+0) {(a)};
        \node[anchor=south west,inner sep=0] at (#1+3.0,#2+0) {\includegraphics[clip, height=\heightsize]{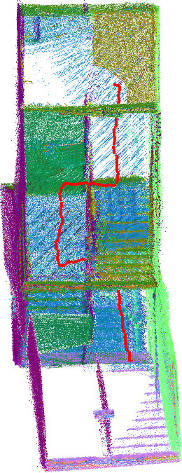}};
        \filldraw[black!70] (#1+0.96,#2+3.05) circle (2.0pt);
        \fill[black!50] (#1+2.27,#2+0.80) rectangle (#1+2.33,#2+1.00);
        \filldraw[black!70] (#1+3.93,#2+3.05) circle (2.0pt);
        \fill[black!50] (#1+3.92,#2+0.87) rectangle (#1+4.12,#2+0.93);
        \draw[black, ultra thick] (#1+2.4,#2+0) rectangle (#1+2.8,#2+3);
        \draw[>=triangle 45, <->] (#1+4.9,#2+0) -- node[rotate=90,anchor=south] {$9.25\,\mathrm{m}$} (#1+4.9,#2+\heightdigit);
    }
    \newcommand\inllama[2]{
        \node[anchor=south west,inner sep=0] at (#1,#2) {\includegraphics[clip, height=\heightsize]{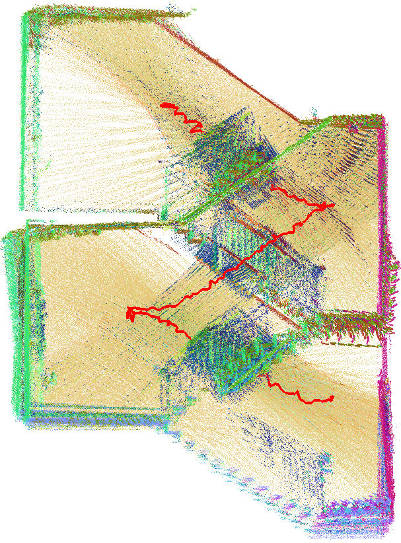}};
        \node[anchor=south,inner sep=0] at (#1+3.65,#2+0) {(b)};
        \node[anchor=south west,inner sep=0] at (#1+3.0,#2+0) {\includegraphics[clip, height=\heightsize]{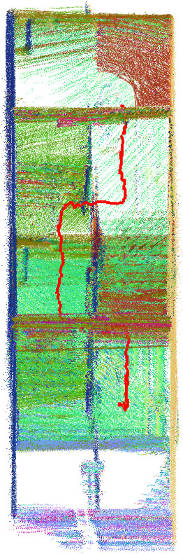}};
        \filldraw[black!70] (#1+1.05,#2+3.02) circle (2.0pt);
        \fill[black!50] (#1+2.27,#2+0.88) rectangle (#1+2.33,#2+1.08);
        \filldraw[black!70] (#1+3.84,#2+3.08) circle (2.0pt);
        \fill[black!50] (#1+3.75,#2+0.95) rectangle (#1+3.95,#2+1.01);
        \draw[black, ultra thick] (#1+2.4,#2+0) rectangle (#1+2.8,#2+3);
    }
    \newcommand\legend[2]{
        \filldraw[black!70] (#1+0.1,#2+0) circle (2.0pt);
        \fill [black!50] (#1+0,#2+0.25) rectangle (#1+0.2,#2+0.31);
        \node[anchor=west, color=black!70, text width=2.7cm, execute at begin node=\setlength{\baselineskip}{8pt}] at (#1+0.22,#2+0.28) {\scriptsize Start position};
        \node[anchor=west, color=black!70, text width=2.7cm, execute at begin node=\setlength{\baselineskip}{8pt}] at (#1+0.22,#2+0) {\scriptsize End position};
    }
    \begin{tikzpicture}
        \node[anchor=south west,inner sep=0] at (0,0) {\includegraphics[clip,trim=0.5cm 0.5cm 1cm 0.65cm, height=4.9cm]{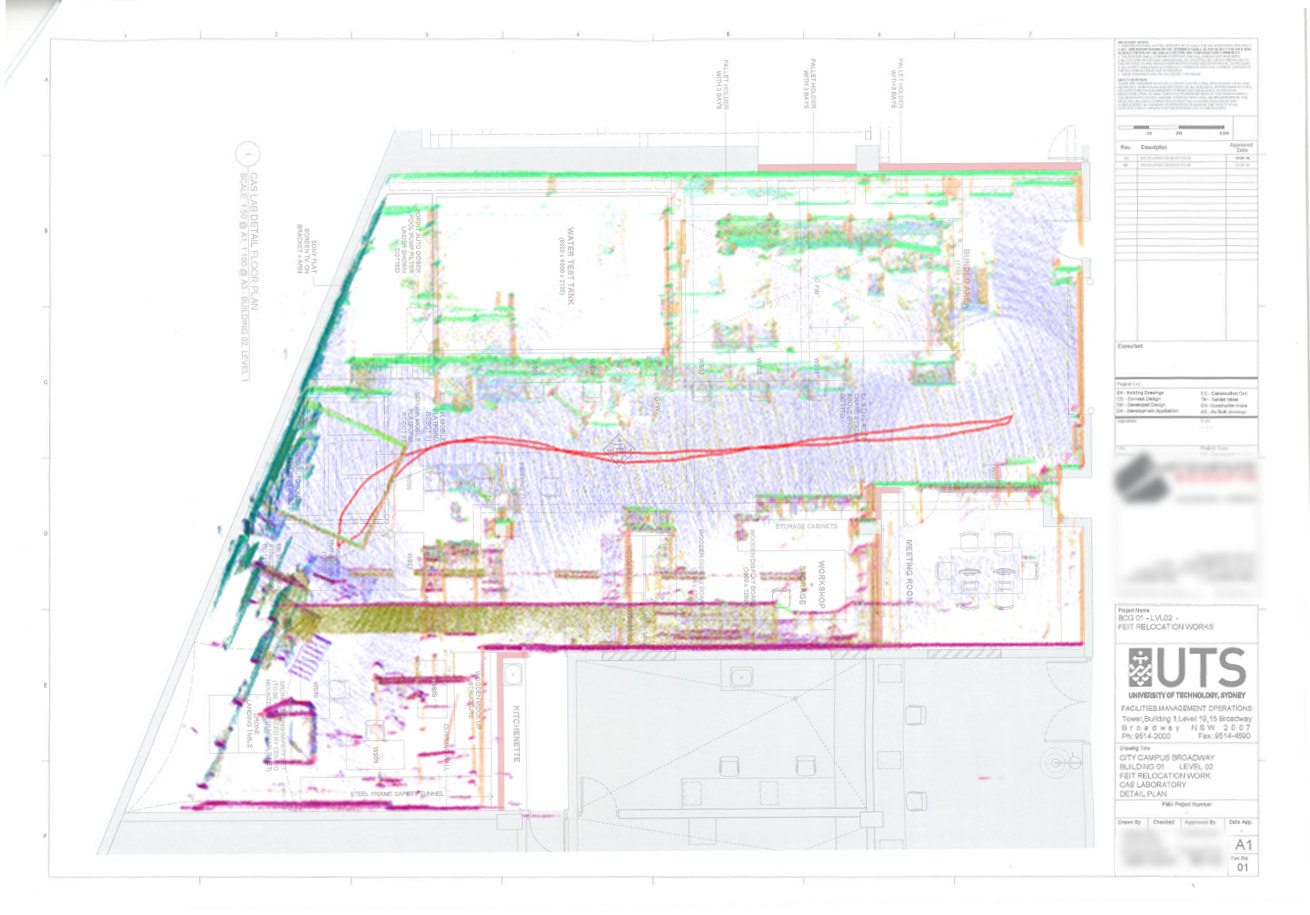}};
        \draw[>=triangle 45, <->] (0.8,0.50) -- node[rotate=90,anchor=south] {$16.5\,\mathrm{m}$} (0.8,4.05);
        \draw[>=triangle 45, <->] (0.85,0.2) -- node[anchor=south] {$24.5\,\mathrm{m}$} (6.05,0.2);
        \node[anchor=south, gray, rotate=67] at (1.49,2.70) {\scriptsize $Wall\ 1$};
        \draw[black, ultra thick, rotate around={67:(1.55,2.40)}] (1.55,2.40) rectangle (2.05,2.60);
        \node[anchor=south, gray] at (5.60,4.2) {\scriptsize $Wall\ 2$};
        \draw[black, ultra thick] (5.35,4.00) rectangle (5.85,4.20);
        \fill[black!50] (1.7,1.95) rectangle (1.9,2.01);
        \filldraw[black!70] (1.85,1.99) circle (2.0pt);
        \loam{8.5}{0.1}
        \inllama{13.5}{0.1}
        \legend{8.25}{0.25}
        \node[anchor=south] at (3.90,4.4) {$(a)$};
        \node[anchor=south] at (10.80,3.25) {$(b)$};
        \node[anchor=south] at (15.80,3.25) {$(c)$};
        \node[anchor=north east, align=right, text width=11cm, execute at begin node=\setlength{\baselineskip}{8pt}] at (17.8,4.95) {\footnotesize\bf{Indoor trajectories and maps}};
        \node[anchor=north east, align=right, text width=11cm, execute at begin node=\setlength{\baselineskip}{8pt}] at (17.8,4.65) {
            \scriptsize 
            coloured with post-computed normals (100 closest neighbours)};
        \node[anchor=north east, align=right, text width=11cm, execute at begin node=\setlength{\baselineskip}{8pt}] at (17.8,4.40) {
            \scriptsize 
            $(a)$ Lab environment with IN2LAAMA, $(b)$ Staircase with \cite{Zhang2014b}, $(c)$ Staircase with \cite{LeGentil2019}};
    \end{tikzpicture}
    \caption{Maps and 6-DoF trajectories estimated in indoor environments. $(a)$ is overlaid over the digitalised pre-construction blueprint (which does not capture reality accurately due to structural differences with the original plans and furniture/equipment present in the lab). The black rectangles highlight the walls used for the quantitative comparison. The staircase map generated by IN2LAAMA is shown in Fig.~\ref{figure:teaser}. Note that \cite{Zhang2014b} is real-time while \cite{LeGentil2019} and IN2LAAMA are offline estimation frameworks.}
    \label{figure:staircase}
\end{figure*}
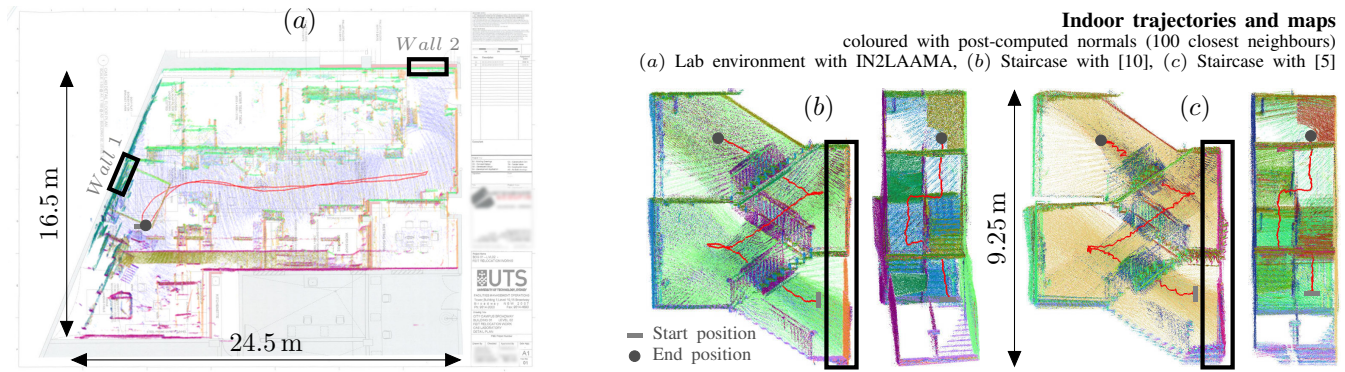

\begin{table}
    \caption{Quantitative comparison on real data.}
    \begin{center}
    \vspace{-0.2cm}
    \newcolumntype{C}[1]{>{\centering\let\newline\\\arraybackslash\hspace{0pt}}m{#1}}
    \def\colsize{1.5cm}
    \def\textsize{\scriptsize}
        \begin{tabular}{C{2cm}|  C{\colsize} C{\colsize}  C{\colsize}}
            \textsize Dataset &\multicolumn{3}{c}{\textsize RMS point-to-plane distance (mm) }\\
            (plane used) &
            \textsize \cite{Zhang2014b}  &
            \textsize \cite{LeGentil2019} &
            \textsize IN2LAAMA
            \\
            \hline
            \rowcolor[gray]{0.95}
            \textsize Lab (floor) & \textsize 24 & \textsize 19 &\textsize \textbf{16}
            \\
            \hline
            \textsize Lab (wall 1) & \textsize 20 & \textsize 13 & \textsize \textbf{11}
            \\
            \hline
            \rowcolor[gray]{0.95}
            \textsize Lab (wall 2) & \textsize 16 & \textsize \textbf{15} & \textsize 16
            \\
            \hline
            \textsize Staircase (wall) & \textsize 60 & \textsize 31 & \textsize \textbf{10}
        \end{tabular}
    \end{center}
    \vspace{-0.1cm}
    {\footnotesize RMS point-to-plane distances between map points and the corresponding plane. Note that the IN2LAAMA and \cite{LeGentil2019} are offline frameworks, whereas \cite{Zhang2014b} operates in real-time.}
    \label{table_real_quantitative}
\end{table}

Fig.~\ref{figure:staircase}\,(a) shows the lab environment map estimated by IN2LAAMA, and Table~\ref{table_real_quantitative} (first three rows) shows the associated quantitative comparison with the other methods.
The metric used is the Root-Mean-Square (RMS) point-to-plane distance between the 3D-lidar points belonging to a plane and the corresponding plane.
The planes are estimated with a principal component analysis on the manually segmented points.
The estimated trajectory is $40.4\,\mathrm{m}$-long, with an average velocity of $1.02\,\mathrm{m/s}$.
Without loop-closure, the proposed method accumulates a drift of $0.11\,\mathrm{m}$ and $0.61\,\mathrm{{}^\circ}$.
As the motion in this dataset is not aggressive, the use of IMU factors does not impact the final estimate significantly.
All the benchmarked methods perform similarly and lead to RMS distances inferior to the lidar noise specification ($\pm 3\,\mathrm{cm}$).

The staircase dataset is more challenging because of the nature of the motion (dynamic with strong rotations) and the weak geometric information contained in some of the collected lidar scans.
The maps generated with \cite{Zhang2014b} and \cite{LeGentil2019} are displayed in Fig.~\ref{figure:staircase} and the one generated with the proposed method in Fig.~\ref{figure:teaser}.
Irrespectively, whether it is only to help motion distortion correction as in \cite{LeGentil2019} or to also fully constrain the pose-graph optimisation (IN2LAAMA approach), tightly integrating inertial information in the trajectory estimation leads to greater mapping accuracy than lidar-only techniques. The constant velocity motion assumption used in \cite{Zhang2014b} reaches its limits in this kind of scenarios.
To provide a quantitative evaluation of the maps, point-to-plane distances are computed for points belonging to the same wall across the different floor levels (black rectangles in Fig.~\ref{figure:staircase}).
The results are shown in the last row of Table~\ref{table_real_quantitative}.

Note that our framework uses extra information (IMU readings) in comparison to \cite{Zhang2014b}, and the incrementally built batch optimisation is not running in real-time, thus, the comparison is not totally fair.
Table~\ref{table:timing} shows the computation time and memory usage for the real-data experiments.
We differentiate between the amount of memory used by the non-linear least-square optimiser and the rest of the memory consumption (data, UPMs, program, etc.).
While Ceres provides smart implementations of commonly used iterative solvers, its design is not optimised memory-wise for very large numbers of residuals as the evaluation of the system's Hessian matrix requires considerable memory allocations.
A potential solution to greatly reduce the memory load is to directly compute the Hessian matrix as done in [31].
With further engineering efforts, the execution time can be substantially reduced as well; parallelisation on GPU could be applied to different operations (e.g. feature extraction, inertial data upsampling, residual and Jacobian computations, etc.).
Note that the proposed framework introduces a principled and accurate full-batch optimisation that leverages the full dataset at the cost of computation time and memory usage.
Nonetheless, we believe that, at the expense of accuracy, simple approximations of our framework associated with the aforementioned effective implementation techniques would enable real-time and efficient localisation and mapping methods.

\begin{table}
    \caption{Memory consumption and execution time.}
    \begin{center}
    \vspace{-0.2cm}
    \newcolumntype{C}[1]{>{\centering\let\newline\\\arraybackslash\hspace{0pt}}m{#1}}
    \def\colsize{1.2cm}
    \def\textsize{\scriptsize}
        \begin{tabular}{ C{2cm}|  C{0.8cm}  C{\colsize} C{\colsize} C{\colsize}}
            \textsize Dataset (length) &
            \textsize $\nevery$ &
            \textsize Mem. data/prog. &
            \textsize Mem. optimiser &
            \textsize Exec. time
            \\
            \hline
            \rowcolor[gray]{0.95}
            \textsize Lab ($41\,\mathrm{s}$) &
            \textsize 100 &
            \textsize $4.70\,\mathrm{GiB}$ &
            \textsize $0.59\,\mathrm{GiB}$ &
            \textsize $950\,\mathrm{s}$
            \\
            \hline
            \textsize Staircase ($35\,\mathrm{s}$) &
            \textsize 10 &
            \textsize $4.20\,\mathrm{GiB}$ &
            \textsize $0.58\,\mathrm{GiB}$ &
            \textsize $1306\,\mathrm{s}$
            \\
            \hline
            \rowcolor[gray]{0.95}
            \textsize Outdoor ($85\,\mathrm{s}$) &
            \textsize 100 &
            \textsize $6.45\,\mathrm{GiB}$ &
            \textsize $5.77\,\mathrm{GiB}$ &
            \textsize $4699\,\mathrm{s}$
        \end{tabular}
    \end{center}
    \vspace{-0.1cm}
    {\footnotesize The outdoor dataset is collected with a Velodyne HDL-32 that produces around four times as much data as the VLP-16 used in the other experiments. For the outdoor dataset, the optional ICP tests to validate/reject loop-closures have been activated. It represents $1588\,\mathrm{s}$ of the overall execution time.}
    \label{table:timing}
\end{table}

\subsubsection{Outdoors}

As mentioned above, the front-end of the proposed method has been designed for structured geometry.
While the geometric features used are largely present in indoor environments, outdoor scenarios can represent a challenge for our feature extraction algorithm.

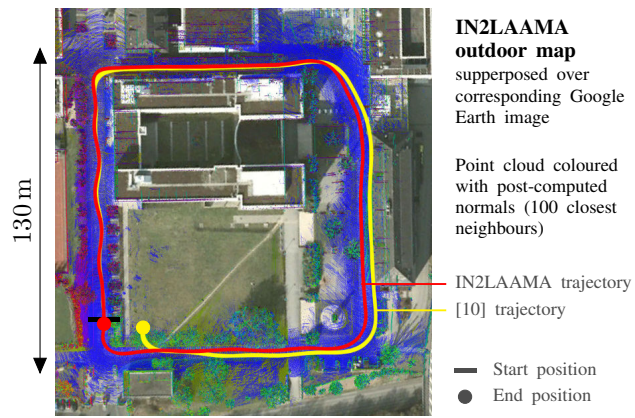
\begin{figure}
    \centering
    \begin{tikzpicture}
        \node[anchor=south west,inner sep=0] at (0,0.04) {\includegraphics[clip, width=5cm, trim= 0cm 0cm 0cm 0cm]{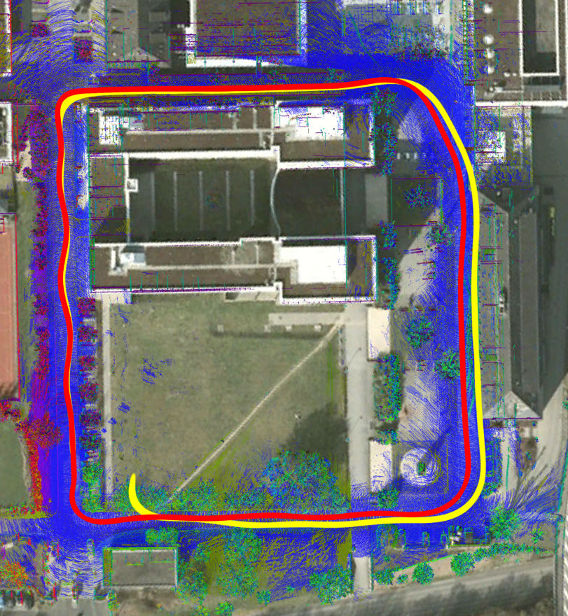}};
        \draw[>=triangle 45, <->] (-0.2,0.6) -- node[rotate=90,anchor=south] {$130\,\mathrm{m}$} (-0.2,4.95);
        \node[anchor=north west, text width=2.4cm, execute at begin node=\setlength{\baselineskip}{9pt}] at (5.2,5.45) {\footnotesize\bf{IN2LAAMA outdoor map}};
        \coordinate (in2laama_traj) at (4.07,1.8);
        \coordinate (loam_traj) at (4.25,1.45);
        \node[anchor=north west, text width=2.7cm, execute at begin node=\setlength{\baselineskip}{8pt}] at (5.2,4.8) {\scriptsize supperposed over corresponding Google Earth image};
        \node[anchor=north west, text width=2.7cm, execute at begin node=\setlength{\baselineskip}{8pt}] at (5.2,3.6) {\scriptsize Point cloud coloured with post-computed normals (100 closest neighbours)};
        \node[anchor=west, color=black!70, text width=2.7cm, execute at begin node=\setlength{\baselineskip}{8pt}] (in2laama_legend) at (5.2,1.8) {\scriptsize IN2LAAMA trajectory};
        \node[anchor=west, color=black!70, text width=2.7cm, execute at begin node=\setlength{\baselineskip}{8pt}] (loam_legend) at (5.2,1.45) {\scriptsize \cite{Zhang2014b} trajectory};
        \node[anchor=west, color=black!70, text width=2.7cm, execute at begin node=\setlength{\baselineskip}{8pt}] at (5.7,0.65) {\scriptsize Start position};
        \node[anchor=west, color=black!70, text width=2.7cm, execute at begin node=\setlength{\baselineskip}{8pt}] at (5.7,0.3) {\scriptsize End position};
        \draw[line width=0.025cm, red] (in2laama_traj) -- (in2laama_legend.west);
        \draw[line width=0.025cm, yellow] (loam_traj) -- (loam_legend.west);
        \fill [black]     (0.44,1.30) rectangle (0.86,1.36);
        \filldraw[yellow] (1.17,1.22) circle (2.5pt);
        \filldraw[red]    (0.65,1.27) circle (2.5pt);
        \filldraw[black!70] (5.45,0.30) circle (2.5pt);
        \fill [black!70] (5.3,0.62) rectangle (5.6,0.68);
    \end{tikzpicture}
    \caption{Map and trajectory generated by IN2LAAMA in outdoor environments (MC2SLAM dataset \cite{Neuhaus2018}, sequence ``campus\_drive").}
    \label{figure:mc2slam}
\end{figure}

We have chosen the MC2SLAM dataset \cite{Neuhaus2018} to show the performance of the proposed approach in an outdoor environment as it provides per-lidar-point timestamps.
The data have been acquired by a \textit{Velodyne HDL-32} lidar and its built-in IMU mounted on top of a car that is driven around a University campus (sequence ``campus\_drive" of \cite{Neuhaus2018}).
Fig.~\ref{figure:mc2slam} shows the map generated by IN2LAAMA with loop-closure.
As no ground truth is given with the dataset, we overlay the map onto the corresponding Google Earth\footnote{https://www.google.com/earth/} image.
The estimated trajectory is $409\,\mathrm{m}$ long and lasts $85.4\,\mathrm{s}$.
Without loop closure, the accumulated drift of the proposed method is $2.45\,\mathrm{m}$ (mostly on the vertical axis: $2.34\,\mathrm{m}$), and $1.12\,{}^\circ$.
The proposed method outperforms \cite{Zhang2014b} that accumulates a significant drift of $9.78\,\mathrm{m}$ and $33.4\,\mathrm{{}^\circ}$ across the recording.

At the start of the dataset, due to the translation-only trajectory, the accelerometer biases are not observable; the car is driven in a straight line for few meters (before starting a series of turns).
Without any additional constraint on the accelerometer biases (Section~VI-D) and given wrong initial orientation ($\imurot^{\frametime_0}$ arbitrarily flipped up-side-down), the estimated state converges toward wrong biases values (magnitude of $2g$).
The extra factor on $\bacccorrectionz$ provides the constraint required to contain the estimation error on the accelerometer biases while the lack of motion variations prevents the estimation from converging toward the true value of the state.

\subsection{Real-data - Calibration}

Finally, this set-up aims to evaluate the accuracy of the extrinsic calibration performed by the proposed framework.
To do so, we benchmark our method against a ``chained calibration" using an extra sensor, an RGB camera (Intel Realsense D435).
The ``chained calibration" computes IMU-camera and camera-lidar extrinsic calibrations and compounds them together to obtain the IMU-lidar geometric transformation.
The intrinsic camera calibration has been performed with \textit{RADOCC} \cite{Kassir2010}.
The IMU-camera geometric transformation has been estimated with \textit{Kalibr} \cite{Furgale2013} by moving the sensor suite in front of a static calibration pattern.
The camera-lidar extrinsic calibration is the result of the minimisation of point-to-plane distances of lidar points belonging to a checkerboard itself characterised by plane equations in the camera frame.
The data used to estimate this last ``link" are static to avoid the issues of motion distortion and time synchronisation.

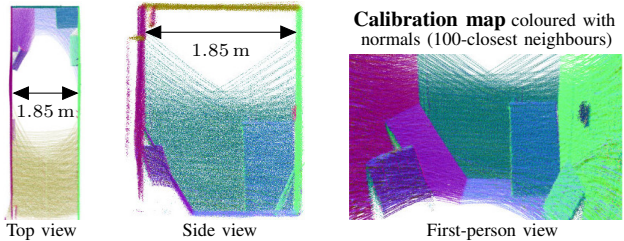
\begin{figure}
    \centering
    \begin{tikzpicture}
        \def\imageheight{2.9}
        \node[anchor=south west,inner sep=0] at (0,0) {\includegraphics[clip, height=\imageheight cm]{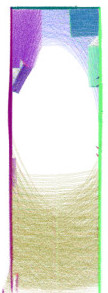}};
        \node[anchor=south west,inner sep=0] at (1.6,0) {\includegraphics[clip, height=\imageheight cm]{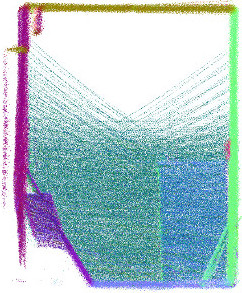}};
        \node[anchor=south west,inner sep=0] at (4.6,0) {\includegraphics[clip, height=2.2 cm]{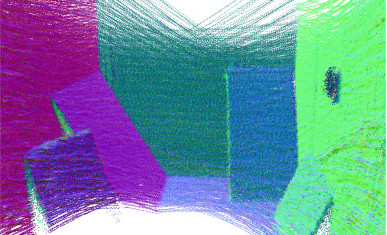}};
        \draw[>=triangle 45, <->] (0.12,1.69) -- (1.0,1.69);
        \node[anchor=north] at (0.56,1.66) {\scriptsize$1.85\,\mathrm{m}$};
        \draw[>=triangle 45, <->] (1.88,2.5) -- (3.90,2.5);
        \node[anchor=north] at (2.89,2.47) {\scriptsize$1.85\,\mathrm{m}$};
        \node[anchor=south west] at (4.53,2.4) {\footnotesize \bf{Calibration map}};
        \node[anchor=south west] at (6.63,2.45) {\scriptsize coloured with};
        \node[anchor=south west] at (4.63,2.15) {\scriptsize normals (100-closest neighbours)};
        \node[anchor=north east] at (1.1, 0.08) {\scriptsize Top view};
        \node[anchor=north east] at (3.5 ,0.08) {\scriptsize Side view};
        \node[anchor=north east] at (7.5 ,0.08) {\scriptsize First-person view};
    \end{tikzpicture}
    \caption{Map generated by IN2LAAMA during the system's calibration.}
    \label{figure:calib_map}
\end{figure}

The two calibration pipelines are executed independently.
Fig.~\ref{figure:calib_map} displays the map estimated during the IN2LAAMA calibration procedure.
Then, the evaluation is conducted by running the proposed method for localisation and mapping (on another dataset) based on the calibration parameters obtained from the two calibration pipelines.
The aggressive trajectory (RMS linear and angular velocities of $0.24\,\mathrm{m/s}$ and $81.3\,\mathrm{{}^\circ/s}$) emphasises the need for good calibration to estimate an accurate map.
In the resulting maps, average (over six non-coplanar planes) RMS point-to-plane distances are computed between 3D points and their associated planes in the scene.
The chained calibration leads to a mean of $58\,\mathrm{mm}$ while IN2LAAMA's map displayed more crispness with an average distance of $27\,\mathrm{mm}$.

The calibration accuracy depends on the quality of the IMU readings as well as the environment and trajectory used for calibration.
With IN2LAAMA's front-end being built upon planar and edge features, the trajectory needs to allow for the frame-to-frame registration of at least three non-coplanar planes or non-collinear edges to constrain the lidar pose estimation properly.
As demonstrated in \cite{Yang2019}, the calibration parameters are not observable for every trajectory.
The ideal ones are random paths that stimulate the IMU's six DoFs.

\section{Conclusion}

This paper introduced \emph{INertial Lidar Localisation Autocalibration And MApping}; a probabilistic framework for lidar-inertial localisation, mapping, and extrinsic calibration.
The proposed method aims to deal with the motion distortion present in lidar scans without the need for an explicit motion model.
The key idea is to use Upsampled Preintegrated Measurements to allow precise characterisation of the system's motion during each lidar scan.
The frame-to-frame scan registration is performed with a full batch on-manifold optimisation based on point-to-plane, point-to-line, and inertial residuals.
The integration of IMU factors allows us to add robustness to highly dynamic motion.
Extensive experiments have been conducted to demonstrate the performances of IN2LAAMA both on simulated and real-world data.
A comparison with the state-of-the-art lidar localisation and mapping algorithm shows that our method performs better across diverse environments.
While providing more accurate results, the current implementation of IN2LAAMA does not allow real-time operations.

Many applications (e.g. 3D mapping as a service) can leverage such computational intensive framework as it is.
However, the proposed method has the potential to be the baseline to develop more efficient lidar-inertial estimation frameworks for real-time operations.
To this end, one can easily think about mechanisms such as local maps \cite{Bosse2003}, sliding window optimisation \cite{Yang2017}, graph sparsification with marginalisation, parallel computation on GPU, etc.
Therefore, future work includes the exploration of different strategies and simplifications to make the method computationally efficient.
Another area for future work is the integration of a more robust and efficient loop closure detection mechanism.
Alternative map representations such as using surfels could also be investigated to integrate frame-to-model constraints in the optimisation and improve the front-end robustness in unstructured environments.

\appendix[Rejection algorithm for lidar feature candidates]
\noindent\hrulefill
    \begin{algorithmic}[1]
    \vspace{-0.1cm}
        \begin{scriptsize}
        \renewcommand{\algorithmicrequire}{\textbf{Input:}}
        \renewcommand{\algorithmicensure}{\textbf{Output:}}
        \Require
            \Statex $\thresholdregressionerror, \thresholdmaxregressionerror$: Thresholds on mean and max regression error
            \Statex $\regressionerrorlefti, \regressionerrorrighti, \maxregressionerrorlefti, \maxregressionerrorrighti$: Regression errors in $\leftseti$ and $\rightseti$ 
        \Ensure
            \Statex Boolean flag: $Accept\_point$ or $Reject\_point$
        \Statex
            \Statex \textbf{function} \Call{RegressionOK}{$\bullet$ = $\leftseti$ or $\rightseti$}
            \Statex \ \ \ \ \Return $(\regressionerrori_{\bullet}\!<\!\thresholdregressionerror )$ \& $(\maxregressionerrori_{\bullet}\! <\! \thresholdmaxregressionerror)$
            \Statex \textbf{end function}
        \Statex
        \Statex \textbf{function} \Call{Occlusion}{$\bullet$ = $\leftseti$ or $\rightseti$}
            \Statex \ \ \ \ \Return $(\regressionslope_{\bullet} \localpointx^{-1}\!+\!\regressionintercept_{\bullet}\!<\!\lvert \lastlidarpointi \lvert)$
            \Statex \textbf{end function}
        \Statex
        
            \If {\Call{RegressionOK}{$\leftseti$} \& \Call{RegressionOK}{$\rightseti$} }
            \State \Return $Accept\_point$
            \ElsIf {!(\Call{RegressionOK}{$\leftseti$}) \& !(\Call{RegressionOK}{$\rightseti$})\\ \hspace{1.15em}}
            \Return $Reject\_point$
        \Else
            \If {\Call{RegressionOK}{$\leftseti$}}
                \State Remove $\lidarpointi$ from $\rightseti$ and recompute regression
                \If { !(\Call{RegressionOK}{$\rightseti$}) $\parallel$ \Call{Occlusion}{$\rightseti$} }
                    \State \Return $Reject\_point$
                \Else
                    \State $\regressionslope_{\rightseti}\!\gets\!(\localpointy^1\!-\!\localpointy^0) / (\localpointx^1\!-\!\localpointx^0)$, recompute $\directionvector_{\rightseti}$
                    \State \Return $Accept\_point$
                \EndIf
            \ElsIf {\Call{RegressionOK}{$\rightseti$}}
                \State Remove $\lidarpointi$ from $\leftseti$ and recompute regression
                \If { !(\Call{RegressionOK}{$\leftseti$}) $\parallel$ \Call{Occlusion}{$\leftseti$} }
                    \State \Return $Reject\_point$
                \Else
                    \State $\regressionslope_{\leftseti}\!\gets\!(\localpointy^0\!-\!\localpointy^{-\!1}) / (\localpointx^0\!-\!\localpointx^{-\!1})$, recompute $\directionvector_{\leftseti}$
                    \State \Return $Accept\_point$
                \EndIf
            \EndIf
        \EndIf

        \end{scriptsize}
    \end{algorithmic}
    \vspace{-0.4cm}
    \hrulefill

\ifCLASSOPTIONcaptionsoff
  \newpage
\fi

\bibliographystyle{IEEEtran}
\bibliography{bare_jrnl}

\begin{thebibliography}{10}
\providecommand{\url}[1]{#1}
\csname url@samestyle\endcsname
\providecommand{\newblock}{\relax}
\providecommand{\bibinfo}[2]{#2}
\providecommand{\BIBentrySTDinterwordspacing}{\spaceskip=0pt\relax}
\providecommand{\BIBentryALTinterwordstretchfactor}{4}
\providecommand{\BIBentryALTinterwordspacing}{\spaceskip=\fontdimen2\font plus
\BIBentryALTinterwordstretchfactor\fontdimen3\font minus
  \fontdimen4\font\relax}
\providecommand{\BIBforeignlanguage}[2]{{%
\expandafter\ifx\csname l@#1\endcsname\relax
\typeout{** WARNING: IEEEtran.bst: No hyphenation pattern has been}%
\typeout{** loaded for the language `#1'. Using the pattern for}%
\typeout{** the default language instead.}%
\else
\language=\csname l@#1\endcsname
\fi
#2}}
\providecommand{\BIBdecl}{\relax}
\BIBdecl

\bibitem{DurrantWhyte1996}
H.~F. Durrant-Whyte, ``{An autonomous guided vehicle for cargo handling
  applications},'' \emph{International Journal of Robotics Research}, vol.~15,
  no.~5, pp. 407--440, 1996.

\bibitem{LeGentil2018}
C.~{Le Gentil}, T.~Vidal-Calleja, and S.~Huang, ``{3D Lidar-IMU Calibration
  based on Upsampled Preintegrated Measurements for Motion Distortion
  Correction},'' \emph{IEEE International Conference on Robotics and
  Automation}, 2018.

\bibitem{Lupton2012}
T.~Lupton and S.~Sukkarieh, ``{Visual-inertial-aided navigation for
  high-dynamic motion in built environments without initial conditions},''
  \emph{IEEE Transactions on Robotics}, vol.~28, no.~1, pp. 61--76, 2012.

\bibitem{Forster2015a}
C.~Forster, L.~Carlone, F.~Dellaert, and D.~Scaramuzza, ``{IMU preintegration
  on manifold for efficient visual-inertial maximum-a-posteriori estimation},''
  \emph{Robotics: Science and Systems}, pp. 6--15, 2015.

\bibitem{LeGentil2019}
C.~{Le Gentil}, T.~Vidal-Calleja, and S.~Huang, ``{IN2LAMA : INertial Lidar
  Localisation And MApping},'' \emph{IEEE International Conference on Robotics
  and Automation}, 2019.

\bibitem{Besl1992}
P.~J. Besl and N.~D. McKay, ``{A Method for Registration of 3-D Shapes},''
  \emph{IEEE Transactions on Pattern Analysis and Machine Intelligence}, 1992.

\bibitem{Segal2009}
A.~Segal, D.~Haehnel, and S.~Thrun, ``{Generalized-ICP},'' \emph{Robotics:
  Science and Systems}, vol.~5, pp. 168--176, 2009.

\bibitem{Mendes2016}
E.~Mendes, P.~Koch, and S.~Lacroix, ``{ICP-based pose-graph SLAM},'' \emph{IEEE
  International Symposium on Safety, Security and Rescue Robotics}, pp.
  195--200, 2016.

\bibitem{Hong2010}
S.~Hong, H.~Ko, and J.~Kim, ``{VICP: Velocity updating iterative closest point
  algorithm},'' \emph{Proceedings - IEEE International Conference on Robotics
  and Automation}, no. Section 3, pp. 1893--1898, 2010.

\bibitem{Zhang2014b}
J.~Zhang and S.~Singh, ``{LOAM : Lidar odometry and mapping in real-time},''
  \emph{Robotics: Science and Systems}, pp. 7--15, 2014.

\bibitem{Bosse2009}
M.~Bosse and R.~Zlot, ``{Continuous 3D Scan-Matching with a Spinning 2D
  Laser},'' \emph{IEEE International Conference on Robotics and Automation},
  2009.

\bibitem{Furgale2012}
P.~Furgale, T.~D. Barfoot, and G.~Sibley, ``{Continuous-Time Batch Estimation
  using Temporal Basis Functions},'' \emph{IEEE International Conference on
  Robotics and Automation}, pp. 2088--2095, 2012.

\bibitem{Anderson2015}
S.~Anderson and T.~D. Barfoot, ``{Full STEAM ahead: Exactly sparse Gaussian
  process regression for batch continuous-time trajectory estimation on
  SE(3)},'' \emph{IEEE International Conference on Intelligent Robots and
  Systems}, vol. 2015-Dec, no.~3, pp. 157--164, 2015.

\bibitem{Droeschel2018}
D.~Droeschel and S.~Behnke, ``{Efficient Continuous-time SLAM for 3D
  Lidar-based Online Mapping},'' \emph{IEEE International Conference on
  Robotics and Automation (ICRA)}, no. May, pp. 5000--5007, 2018.

\bibitem{Bosse2012}
M.~Bosse, R.~Zlot, and P.~Flick, ``{Zebedee : Design of a spring-mounted 3-D
  range sensor with application to mobile mapping},'' \emph{IEEE Transactions
  on Robotics}, vol.~28, no. October, pp. 1--15, 2012.

\bibitem{Park2018}
C.~Park, P.~Moghadam, S.~Kim, A.~Elfes, C.~Fookes, and S.~Sridharan, ``{Elastic
  LiDAR Fusion: Dense Map-Centric Continuous-Time SLAM},'' \emph{IEEE
  International Conference on Robotics and Automation}, 2018.

\bibitem{Geneva2018}
P.~Geneva and K.~Eckenhoff, ``{LIPS: LiDAR-Inertial 3D Plane SLAM},''
  \emph{IEEE International Conference on Intelligent Robots and Systems}, 2018.

\bibitem{Serafin2016}
J.~Serafin, E.~Olson, and G.~Grisetti, ``{Fast and robust 3D feature extraction
  from sparse point clouds},'' \emph{IEEE International Conference on
  Intelligent Robots and Systems}, vol. 2016-Nov, pp. 4105--4112, 2016.

\bibitem{Deschaud2018}
J.-E. Deschaud, ``{IMLS-SLAM: scan-to-model matching based on 3D data},''
  \emph{IEEE International Conference on Robotics and Automation (ICRA)}, pp.
  2480--2485, 2018.

\bibitem{Furgale2013}
P.~Furgale, J.~Rehder, and R.~Siegwart, ``{Unified temporal and spatial
  calibration for multi-sensor systems},'' \emph{IEEE International Conference
  on Intelligent Robots and Systems}, pp. 1280--1286, 2013.

\bibitem{Taylor2014}
Z.~Taylor and J.~Nieto, ``{Parameterless automatic extrinsic calibration of
  vehicle mounted lidar-camera systems},'' \emph{International Conference on
  Robotics and Automation: Long Term Autonomy Workshop}, no. October, pp. 3--6,
  2014.

\bibitem{Castorena2016}
J.~Castorena, U.~S. Kamilov, and P.~T. Boufounos, ``{Autocalibration of LIDAR
  and optical cameras via edge alignement},'' \emph{IEEE International
  Conference on Acoustics, Speech and Signal Processing (ICASSP)}, pp.
  2862--2866, 2016.

\bibitem{Rasmussen2006}
C.~E. Rasmussen and C.~K.~I. Williams, \emph{{Gaussian Processes for Machine
  Learning}}.\hskip 1em plus 0.5em minus 0.4em\relax The MIT Press, 2006.

\bibitem{Bentley1975}
J.~L. Bentley, ``{Multidimensional binary search trees used for associative
  searching},'' \emph{Communications of the ACM}, vol.~18, no.~9, pp. 509--517,
  sep 1975.

\bibitem{Tereshkov2013}
V.~M. Tereshkov, ``{An Intuitive Approach to Inertial Sensor Bias
  Estimation},'' \emph{International Journal of Navigation and Observation},
  vol. 2013, pp. 1--6, 2013.

\bibitem{Yu2016}
Z.~Yu and J.~L. Crassidis, ``{Accelerometer Bias Calibration Using Attitude and
  Angular Velocity Information},'' \emph{Journal of Guidance, Control, and
  Dynamics}, vol.~39, no.~4, pp. 741--753, 2016.

\bibitem{Du2017}
S.~Du, W.~Sun, and Y.~Gao, ``{Improving observability of an inertial system by
  rotary motions of an IMU},'' \emph{Sensors (Switzerland)}, vol.~17, no.~4,
  pp. 1--20, 2017.

\bibitem{Tereshkov2015}
V.~M. Tereshkov, ``{A Simple Observer for Gyro and Accelerometer Biases in Land
  Navigation Systems},'' \emph{Journal of Navigation}, vol.~68, no.~04, pp.
  635--645, 2015.

\bibitem{Geiger2012}
A.~Geiger, P.~Lenz, and R.~Urtasun, ``{Are we ready for autonomous driving? the
  KITTI vision benchmark suite},'' \emph{Proceedings of the IEEE Computer
  Society Conference on Computer Vision and Pattern Recognition}, pp.
  3354--3361, 2012.

\bibitem{Neuhaus2018}
F.~Neuhaus, T.~Ko{\ss}, R.~Kohnen, and D.~Paulus, ``{MC2SLAM: Real-Time
  Inertial Lidar Odometry Using Two-Scan Motion Compensation},'' \emph{German
  Conference on Pattern Recognition}, 2018.

\bibitem{Kassir2010}
A.~Kassir and T.~Peynot, ``{Reliable Automatic Camera-Laser Calibration},''
  \emph{Australasian Conference on Robotics and Automation}, 2010.

\bibitem{Yang2019}
Y.~Yang, P.~Geneva, K.~Eckenhoff, and G.~Huang, ``{Degenerate Motion Analysis
  for Aided INS with Online Spatial and Temporal Sensor Calibration},''
  \emph{IEEE Robotics and Automation Letters}, vol.~4, no.~2, pp. 2070--2077,
  2019.

\bibitem{Bosse2003}
M.~Bosse, P.~Newman, J.~Leonard, M.~Soika, W.~Feiten, and S.~Teller, ``{An
  Atlas framework for scalable mapping},'' \emph{Proceedings - IEEE
  International Conference on Robotics and Automation}, vol.~2, pp. 1899--1906,
  2003.

\bibitem{Yang2017}
Z.~Yang and S.~Shen, ``{Monocular visual-inertial state estimation with online
  initialization and camera-IMU extrinsic calibration},'' \emph{IEEE
  Transactions on Automation Science and Engineering}, vol.~14, no.~1, pp.
  39--51, 2017.

\end{thebibliography}

\begin{IEEEbiography}[{\includegraphics[width=1in,height=1.25in,clip,keepaspectratio]{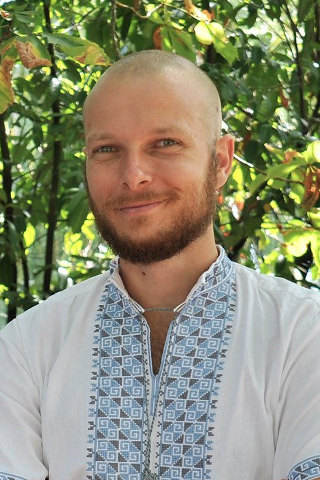}}]{Cedric Le Gentil} received a DUT (associate degree) in electrical engineering and industrial computing from the University Institute of Technology of Cachan (Paris-Sud University) in 2012, and a MSc in electronic and computer sciences from CentraleSupelec in 2015.
Currently working toward a PhD in robotics started in 2017 at the Centre for Autonomous Systems at the University of Technology Sydney, he has recently been on academic visits at the German Aerospace Center (DLR) and the Autonomous Systems Lab, ETHZ, Switzerland.
\end{IEEEbiography}

\begin{IEEEbiography}[{\includegraphics[width=1in,height=1.25in,clip,keepaspectratio]{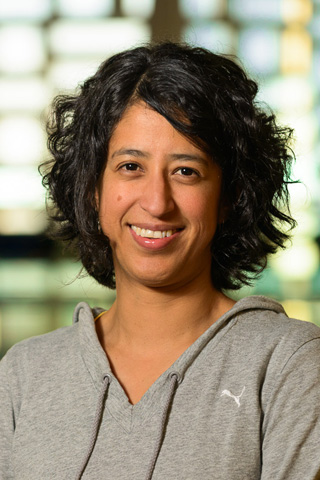}}]{Teresa Vidal-Calleja} received her BSc in Mechanical Engineering from the National Autonomous University of Mexico (UNAM), her MSc in Electrical Engineering (Mechatronics options) from CINVESTAV-IPN, Mexico City, and her PhD in Automatic Control, Computer Vision and Robotics from the Polytechnic University of Catalonia (UPC), Barcelona, Spain in 2007. She was a postdoctoral research fellow at both, LAAS-CNRS in Toulouse, France and the Australian Centre for Field Robotics at the University of Sydney, Australia. She joined the Centre for Autonomous Systems at University of Technology Sydney (UTS) in 2012, where was UTS Chancellor’s Research Fellow and later became Senior Lecturer. She has been visiting scholar at the Active Vision Laboratory of the University of Oxford, UK and, more recently, at the Autonomous Systems Lab, ETHZ, Switzerland. Her research interests are in robotic probabilistic perception.
\end{IEEEbiography}

\begin{IEEEbiography}[{\includegraphics[width=1in,height=1.25in,clip,keepaspectratio]{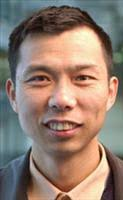}}]{Shoudong Huang} received the Bachelor and Master degrees in Mathematics, Ph.D in Automatic Control from Northeastern University, P.R. China in 1987, 1990, and 1998, respectively.  After his PhD study, he worked at the University of Hong Kong for 1.5 years and The Australian National University for 2 years as a Research Fellow in control area. He joined the Australian Research Council (ARC) Centre of Excellence for Autonomous Systems in 2004 and started to work in robotics area.  He is currently an Associate Professor at Centre for Autonomous Systems, Faculty of Engineering and Information Technology, University of Technology, Sydney, Australia. His research interests include mobile robots simultaneous localization and mapping (SLAM), exploration and navigation and nonlinear system control. He has published more than 150 papers in robotics and control area. He is currently serving as an Associate Editor for IEEE Transactions on Robotics.
\end{IEEEbiography}

\end{document}